\documentclass[sigconf]{acmart}
\usepackage{multirow}
\usepackage{balance}
\usepackage{epstopdf}
\usepackage{graphicx}
\usepackage{makecell}
\usepackage{multirow}

\AtBeginDocument{%
  \providecommand\BibTeX{{%
    \normalfont B\kern-0.5em{\scshape i\kern-0.25em b}\kern-0.8em\TeX}}}

\copyrightyear{2022}
\acmYear{2022}
\setcopyright{acmcopyright}
\acmConference[MM '22] {Proceedings of the 30th ACM International Conference on Multimedia }{October 10--14, 2022}{Lisboa, Portugal.}
\acmBooktitle{Proceedings of the 30th ACM International Conference on Multimedia (MM '22), October 10--14, 2022, Lisboa, Portugal}
\acmPrice{15.00}
\acmISBN{978-1-4503-9203-7/22/10}
\acmDOI{10.1145/3503161.3547873}

\begin{document}

\title{Symmetric Uncertainty-Aware Feature Transmission for \\Depth Super-Resolution}



\author{Wuxuan Shi}
\email{wuxuanshi@whu.edu.cn}
\affiliation{
    \institution{School of Computer Science, Wuhan University}
    \city{Wuhan}
    \country{China}
}

\author{Mang Ye}
\email{yemang@whu.edu.cn}
\authornote{Corresponding Author.}
\affiliation{
    \institution{School of Computer Science, Wuhan University}
    \institution{Hubei Luojia Laboratory}
    \city{Wuhan}
    \country{China}
}

\author{Bo Du}
\email{dubo@whu.edu.cn}
\affiliation{
    \institution{School of Computer Science, Wuhan University}
    \institution{Hubei Luojia Laboratory}
    \city{Wuhan}
    \country{China}
}

\renewcommand{\shortauthors}{Wuxuan Shi, Mang Ye, \& Bo Du}





\begin{abstract}
    Color-guided depth super-resolution (DSR) is an encouraging paradigm that enhances a low-resolution (LR) depth map guided by an extra high-resolution (HR) RGB image from the same scene. Existing methods usually use interpolation to upscale the depth maps before feeding them into the network and transfer the high-frequency information extracted from HR RGB images to guide the reconstruction of depth maps. However, the extracted high-frequency information usually contains textures that are not present in depth maps in the existence of the cross-modality gap, and the noises would be further aggravated by interpolation due to the resolution gap between the RGB and depth images. 
    To tackle these challenges, we propose a novel \textbf{S}ymmetric \textbf{U}ncertainty-aware \textbf{F}eature \textbf{T}ransmission (SUFT) for color-guided DSR. (1) For the resolution gap, SUFT builds an iterative up-and-down sampling pipeline, which makes depth features and RGB features spatially consistent while suppressing noise amplification and blurring by replacing common interpolated pre-upsampling. (2) For the cross-modality gap, we propose a novel Symmetric Uncertainty scheme to remove parts of RGB information harmful to the recovery of HR depth maps. Extensive experiments on benchmark datasets and challenging real-world settings suggest that our method achieves superior performance compared to state-of-the-art methods. Our code and models are available at \textcolor{red}{\url{https://github.com/ShiWuxuan/SUFT}}.
\end{abstract}
\vspace{-0.3cm}

\ccsdesc[500]{Computing methodologies~Reconstruction}
\ccsdesc[500]{Computing methodologies~3D imaging}

\keywords{Depth map, Super-resolution, Uncertainty}


\maketitle
\vspace{-0.4cm}
\begin{figure}[ht]
  \centering
  \includegraphics[width=\linewidth]{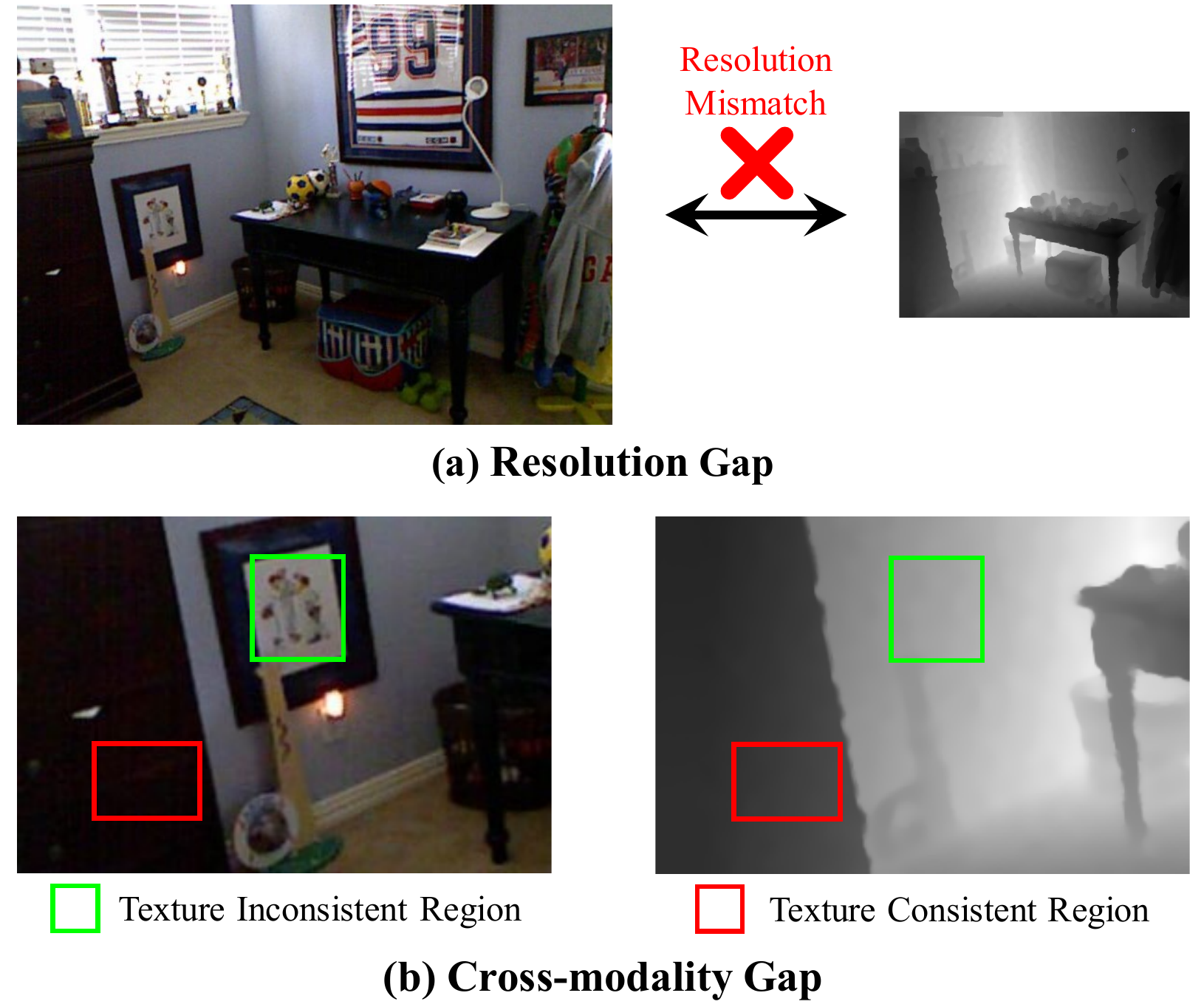}
  \vspace{-0.8cm}
  \caption{Illustration of the two challenges of color-guided DSR. (a) Resolution Gap: The spatial resolution of the RGB guidance image is different from that of the input depth map. (b) Cross-Modality Gap: The images of two modalities have a consistent texture in the red box. However, the green box area is richly textured in the RGB image and the area at the same location in the depth map is smooth.}
  \label{fig:challenge}
  \vspace{-0.6cm}
\end{figure}

\section{Introduction}

\label{Sec 1}
The depth map is an important supplement to the RGB modality, which provides depth information to help the human or computer vision system understand the scene. A better comprehension of the scene is beneficial for research in many areas of computer vision, such as scene recognition \cite{CVPR2016scenerecognition}, autonomous navigation \cite{IROS2013SLAM}, 3D reconstruction \cite{TPAMI20183Dreconstruction}, all of which often require high-quality depth information. Meanwhile, the growing popularity of consumer-grade depth cameras provides great convenience for obtaining depth information. Nevertheless, the resolution of depth maps is difficult to match with RGB images due to the limitation of depth sensor imaging capability, which limits its practical application. As a powerful solution for this issue, depth super-resolution (DSR) technology has attracted increasing attention.

DSR aims to reconstruct an HR depth map from an LR depth map. It needs to recover the lost high-frequency information from the degraded LR depth map while avoiding noise amplification and blurring as much as possible. In practical application scenarios, registered RGB images are frequently used as guidance to improve depth map recovery. Color-guided depth super-resolution is the name for this type of method. The idea behind the approach is that HR RGB images are easy to obtain (consumer-grade depth cameras are typically coupled with RGB sensors), and there is a strong structural similarity between the registered RGB image and the depth map \cite{TIP2017co, SPL2016saliency, TCYB2017iterative}. Color-guided DSR methods have become the dominant solution for the DSR task because they can better recover the detail in the depth map by compensating the lost high-frequency information in the LR depth map with the HR RGB image. Traditional methods \cite{CVPR2013filter1, TPAMI2012filter2, ICCV2013TGV, ICCV2011Park} based on hand-crafted filters or optimization may transmit erroneous structures to the target depth map \cite{ECCV2016DJF}, which causes the recovered depth map to be inaccurate.

With the rapid development of deep learning techniques in recent years, many learning-based color-guided DSR methods \cite{TIP2018CCFN, PR2019CDcGAN, ICCV2109PixTrasnform, ICCV2019GbFT, TCSVT2019MFR, NC2022PDRNet} have been proposed. Although existing algorithms have achieved impressive performance, there are still some unsatisfactory aspects because of two challenges. Firstly, the resolution gap between the paired HR RGB image and LR depth map, as shown in Figure \textcolor{red}{\ref{fig:challenge}} (a), leads to the difficulty of directly fusing RGB features with depth features without any upsampling operation on the original depth map. Because of this challenge, it is impractical to directly apply the post-upsampling \cite{TPAMI2020survey} structure commonly used in single image super-resolution methods. Previous works \cite{TIP2018DepthSRNet, TPAMI2019DJFR, PR2020DSRN, SPL2021DEAFNet} would typically start by using interpolation (\textit{e.g.}, bicubic interpolation) to upscale the depth map to the same resolution as the registered RGB image. However, this strategy frequently introduces side effects (\textit{e.g.}, noise amplification and blur phenomenon). Therefore, \textit{1) how to make depth features and RGB features consistent in spatial size while avoiding the above side effects} is an important issue. Secondly, the edge and texture of the depth map do not always match the registered RGB image, which is called the cross-modality gap. For example, as illustrated in Figure \textcolor{red}{\ref{fig:challenge}} (b), there is a strong similarity in the texture of the red box region in both the RGB image and the depth map. In contrast, the complex texture in the green boxed region of the RGB image presents inconsistency with the smooth texture in the same position region of the depth map. Existing methods \cite{ECCV2016MSGNet, ECCV2016DJF, CVPR2021FDSR} usually have two branches or sub-networks, one for extracting features of LR depth maps and the other for extracting features of corresponding HR RGB images. To provide guidance to DSR, they concatenate the RGB features or the extracted RGB high-frequency features with the depth features. However, due to the cross-modality gap \cite{ICCV21CA}, not all high-frequency information of the RGB images is required for depth map reconstruction. Such approaches tend to transmit mismatched texture information causing texture-copying artifacts and depth bleeding \cite{TCYB2020textcopy, ECCV2020depthbleed}. Consequently, for color-guided DSR methods, \textit{2) it is crucial to accurately estimate the regions of the RGB features that do not match the depth texture during feature transmission.}

To address the aforementioned challenges, we propose a novel feature transmission paradigm for color-guided DSR, called Symmetric Uncertainty-aware Feature Transmission (SUFT), which is embedded into a multi-stage fusion network. To handle the resolution gap, SUFT builds an iterative up-and-down sampling pipeline to replace the commonly used interpolated pre-upsampling in existing methods. Specifically, we upsample the depth features before each feature fusion to make them spatially consistent with the RGB features and project the HR features back into LR spatial domain after each fusion for subsequent operations. In this way, we can eliminate the resolution gap while providing an error feedback mechanism for projection errors at each fusion stage to alleviate the noise amplification and blur. Although \cite{CVPR2021CTKT} has tried to introduce iterative up-and-down sampling in color-guided DSR method, it embeds DBPN \cite{CVPR2018DBPN} in the DSRNet as a base module, which only processes the depth features and cannot refine the RGB features. It limits the effectiveness in the presence of resolution gap between features of two modalities. Our iterative up-and-down sampling method operates on the \textbf{fused features}, which can improve the overall feature representations. Besides, the upsampling layers with convolutional operations provide a difference between the depth features for later uncertainty calculation.

For the cross-modality gap, previous methods cannot accurately estimate regions of RGB features that do not match the depth texture, resulting in texture-copying artifacts in the reconstruction. We design a novel Symmetric Uncertainty scheme to remedy this deficiency via a simple yet effective flipping operation for the uncertainty estimation in feature maps. The uncertainty in the texture and edge regions is much larger than in the smooth regions, allowing the network to know where in the depth map is difficult for DSR. Although several works have studied the behavior of uncertainty for super-resolution \cite{ICMLA2019GRAM, CVPR2021FastBNN, NuerIPS2021UDL}, they all merely exploit uncertainty to optimize the constraints on the network during training and require an additional stage to generate the uncertainty map. Thus, these solutions cannot process the uncertain RGB features during feature transmission. In comparison, our symmetric uncertainty can be directly embedded into a multi-stage fusion network and generate uncertainty maps by feature flipping before feature transmission. Another advantage of our scheme is that the network only propagates forward once at each iteration during both training and testing, reducing the redundant consumption of computational resources. More discussion on the rationale and advantage of feature-level flipping will be detailed in Section \textcolor{red}{\ref{sec:motivation}}.

To summarize, our major contributions are as follows:
\begin{itemize}
    \item To bridge the resolution gap, we build an iterative up-and-down sampling pipeline in feature transmission instead of interpolated pre-upsampling to suppress noise amplification and blur phenomenon.
    \item To narrow the cross-modality gap, we propose Symmetric Uncertainty, which attempts to precisely select the effective information in RGB images, avoiding inconsistent textures.
    \item Combining the above two ideas, we propose Symmetric Uncertainty-aware Feature Transmission (SUFT). Our network achieves superior performance with clearer margins on benchmark datasets and challenging real-world settings. We also provide ablation experiments to analyze the influence of each component in the model.
\end{itemize}

\begin{figure*}[htbp]
  \centering
  \includegraphics[width=17.2cm]{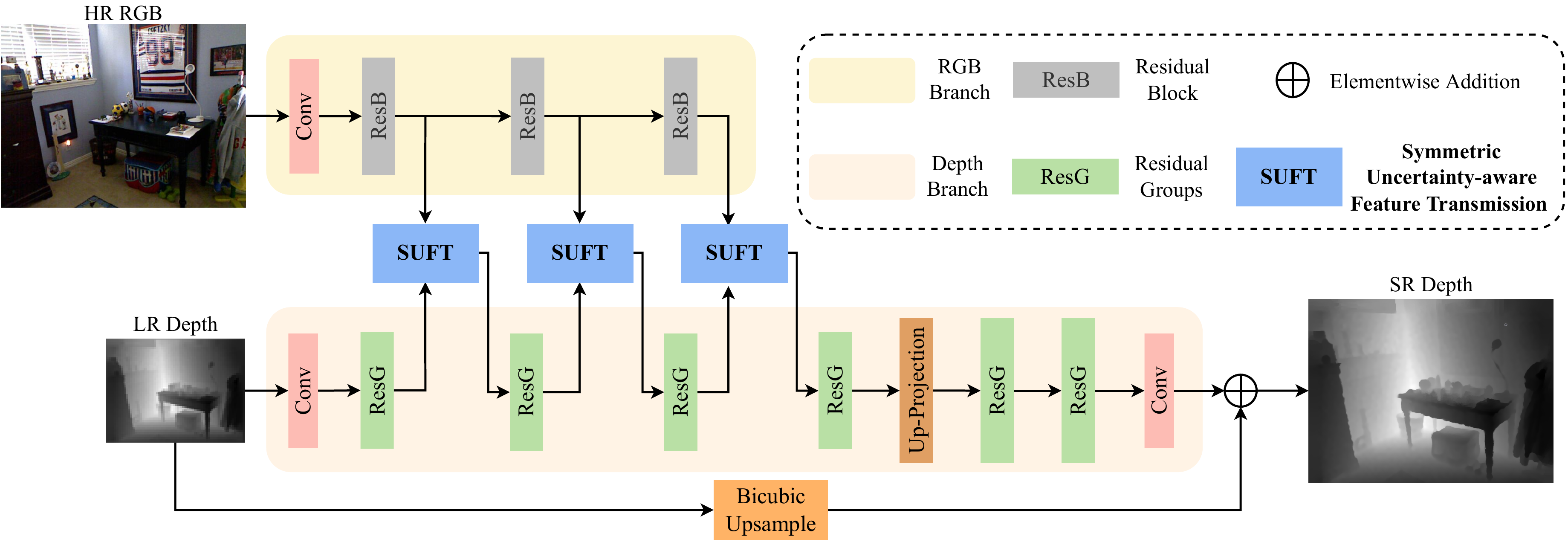}
  \vspace{-0.4cm}
  \caption{Overview of SUFT network architecture. RGB branch extracts features from the RGB image and transfers the features that contribute to depth map recovery to the depth branch by SUFT module processing. The depth branch uses an input depth map and the fused features that have been processed by SUFT modules to generate the high-frequency components of the HR depth map. Finally, the output of depth branch and long skip connection generate the HR depth map by element-wise addition.}
  \label{fig:network}
  \vspace{-0.5cm}
\end{figure*}

\section{Related Work}
\label{Sec 2}
\subsection{Depth Super-Resolution}
Existing methods for solving DSR tasks can be categorized into non color-guided DSR \cite{ECCV2016atgvnet, MM2020SCSN}, where the input data is only an LR depth map, or color-guided DSR \cite{TIP2018DepthSRNet, TIP2020pmbanet}, where the input data includes an LR depth map and a registered HR RGB image. Among the former, Riegler \textit{et al.} \cite{ECCV2016atgvnet} combined a deep convolutional network with a powerful variational model to recover accurate high-resolution depth maps by integrating an anisotropic total generalized variational regularization term at the top of the network. Ye \textit{et al.} \cite{MM2020SCSN} proposed a deep controllable slicing network, which divides regions of different depth ranges to process these regions separately through multiple branches to produce accurate depth recovery.

On the other hand, color-guided methods utilize HR RGB information for better recovery and have become the mainstream of DSR. Guo \textit{et al.} \cite{TIP2018DepthSRNet} built the network based on a residual U-Net to achieve multi-level receptive fields and incorporated multi-level RGB features in the decoding stage. Ye \textit{et al.} \cite{TIP2020pmbanet} proposed a progressive multi-branch aggregation network by using a parallel learning structure and used feedback connections to capture high-frequency features in the depth maps. Beyond the direct fusion of RGB features and depth features, some recent works have started to process the features extracted from RGB images. Tang \textit{et al.} \cite{MM2021bridgenet} proposed a joint learning network of depth super-resolution and monocular depth estimation, considering that features learned from RGB images oriented to the monocular depth estimation task can achieve this cross-modality information transformation in training. Although this is a fascinating work, high-frequency features in monocular depth estimation networks do not always complete the modality transformation accurately, especially in the feature encoding stage. It remains to be explored how to actively and accurately estimate the areas where RGB features and depth textures mismatch.

\begin{figure*}[htbp]
  \centering
  \includegraphics[width=17.5cm]{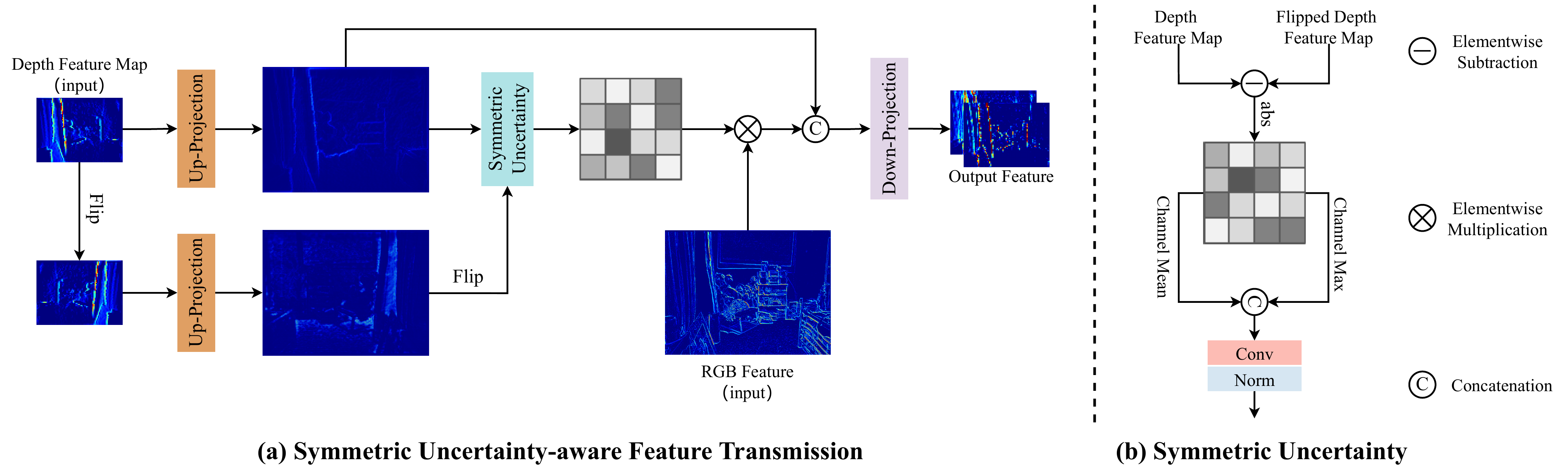}
  \vspace{-0.4cm}
  \caption{Illustration of (a) Symmetric Uncertainty-aware Feature Transmission and (b) Symmetric Uncertainty. We first copy the input features and flip them horizontally. After up-projecting both, we flip the copied features horizontally again to make the two features aligned and input them to the Symmetric Uncertainty module. Symmetric Uncertainty calculates the absolute difference between two features and aggregates channel information by pooling operations. Then it obtains the guidance map by a convolutional layer and normalization to weight the RGB features. After cascading the processed RGB features with the up-projected depth features, the fused features are mapped back into LR domain for subsequent operations in the network.}
  \label{fig:SUFT}
  \vspace{-0.4cm}
\end{figure*}

\vspace{-0.3cm}
\subsection{Uncertainty Modeling}
The nature of uncertainty have been studied in much of the open literature \cite{NuerIPS1997Uncertainty, TNN1999BNN, NuerIPS1996BNN, MVA2019uncertainty, AAAI2022uncertainty, CVPR2020uncertainty}. Uncertainty is increasingly studied and widely used for various vision tasks, such as image classification \cite{CVI2015}, image segmentation \cite{NuerIPS2017Uncertainty, TPAMI2017Segnet}, and depth estimation \cite{NuerIPS2017Uncertainty, CVPR2020UncertaintyMDE}.

To the best of our knowledge, only three works have introduced uncertainty in super-resolution to improve network performance. Lee \textit{et al.} \cite{ICMLA2019GRAM} achieved better results by reducing the loss attenuation of pixels with large uncertainty. Nevertheless, the benefit of this method gradually fails when the pixel variance is high. Kar \textit{et al.} \cite{CVPR2021FastBNN} adopted the uncertainty estimation technique using the batch-normalization \cite{ICML2015BN} layer and explored the adversarial defense mechanism in the image reconstruction domain for the first time. Ning \textit{et al.} \cite{NuerIPS2021UDL} proposed an adaptive weighted loss replacing $MSE$ loss or $L_{1}$ loss to make the network focus on pixels with large uncertainty. While \cite{CVPR2021FastBNN, NuerIPS2021UDL} achieve significant performance gains, additional computational resources are consumed as a consequence. \cite{CVPR2021FastBNN} requires an additional step to estimate the stochastic parameters of each BN layer after training and the same image will be concatenated several times during testing. Although \cite{NuerIPS2021UDL} only changes the loss function, its training process is divided into two stages and the increased computation is visible, especially for super-resolution. Existing methods mainly focus on estimating uncertainty to judge the confidence of the reconstruction results or using uncertainty to optimize the constraints on the network. They cannot generate uncertainty directly during the forward propagation of the network to process RGB features. 

\vspace{-0.3cm}
\subsection{Feature Transmission}
Feature transmission is a key component that affects the performance of color-guided DSR methods. The early methods usually transfer information from RGB features to depth features by element-wise addition \cite{ICCV2109PixTrasnform} or concatenation along the channel axis \cite{ECCV2016DJF, ECCV2016MSGNet}. However, such approaches merge unnecessary information into the depth features as well. 

Recently, some works have attempted to explore more effective ways of feature transmission. He \textit{et al.} \cite{CVPR2021FDSR} designed a high-frequency layer (HFL) by octave convolution \cite{ICCV2019OCConv} to adaptively highlight the high-frequency components extracted from the RGB images during feature transmission to generate the HR depth maps with clearer boundaries. Tang \textit{et al.} \cite{MM2021bridgenet} designed a high-frequency attention bridge (HABdg), which learns the high-frequency attention from the features encoded by the MDENet and weights the original features to obtain the refined high-frequency features. Although these efforts have achieved promising results, they omit that not all high-frequency information in the RGB features is beneficial for depth map reconstruction. 

\section{PROPOSED METHOD}
\subsection{Problem Formulation}
Color-guided DSR aims at recovering the HR depth map $D_{hr} \in \mathbb{R}^{1 \times sH \times sW}$ from the LR depth map $D_{lr} \in \mathbb{R}^{1 \times H \times W}$ with the guidance of corresponding HR RGB image $I_{g} \in \mathbb{R}^{3 \times sH \times sW}$, where $H$, $W$ and $s$ represent the height, width of $D_{lr}$ and the scaling factor, respectively. Generally, a color-guided DSR network is fed with the paired $D_{lr}$ and $I_{g}$, and is trained to generate HR output $D_{sr}$ close to the given target $D_{hr}$. Such process can be expressed as follows: 
\begin{equation} 
D_{hr} \approx  D_{sr} = \mathcal{F}(D_{lr}, \mathcal{E}(I_{g});\theta),
\end{equation}
where $\mathcal{F}$ is the color-guided DSR model to learn the nonlinear mapping from $D_{lr}$ to $D_{hr}$, $\mathcal{E}$ is an extractor to provide guidance information from $I_{g}$, and $\theta$ denotes the learned parameters of $\mathcal{F}$.

For a fair comparison, we apply the same loss function as previous works. Given a training set $\left \{ D_{lr}^i,I_{g}^i,D_{hr}^i \right \}^N_{i=1}$, it contains $N$ LR depth maps with the corresponding HR guidance RGB images as input and target HR depth maps as ground truth. The goal of training our network is to minimize the $L_{1}$ loss function, which has shown improved performance and convergence over $L_{2}$ loss: 
\vspace{-0.1cm}
\begin{equation}
\vspace{-0.1cm}
L(\theta ) = \frac{1}{N}\sum_{i=1}^{N} \left \| \mathcal{F}(D^i_{lr}, I^i_{g})-D^i_{hr} \right \| _1.
\end{equation}
In Section \textcolor{red}{\ref{sec:trainning}}, the training details will be provided.

\vspace{-0.3cm}
\subsection{Network Architecture}
Our proposed SUFT network is shown in Figure \textcolor{red}{\ref{fig:network}}, which consists of two branches (\textit{i.e.}, RGB branch and depth branch) and several SUFT modules. The RGB branch is equipped with three residual blocks to extract information from the RGB image gradually \cite{TPAMI2020yemang}. In the depth branch, we employ Residual Group \cite{ECCV2018RCAN} as the basic module to better mine the information in the fused features. It stacks several simplified residual blocks \cite{CVPRW2017EDSR} with short skip connections and channel attention \cite{ECCV2018CBAM}. The short skip connections make the main network focus on learning high-frequency information and the channel attention allows the depth branch concentrates on useful channels. The first four Residual Groups in the shallow layer contain four residual blocks for each. After the up-projection unit, the two Residual Groups each contain eight residual blocks contemplating the benefits of a deeper network on super-resolution. Then, we introduce the multi-stage feature aggregation strategy to enhance the ability of detail recovery. The long skip connection at the bottom forces the backbone network to focus on learning high-frequency information in DSR. We do not upsample the LR depth map before it is fed into the depth branch. As a result, the spatial size of the depth and RGB features is inconsistent and we can not directly concatenate them. Thus, SUFT can handle the cross-resolution cross-modality feature transmission. 
\vspace{-0.3cm}
\subsection{Symmetric Uncertainty-aware Feature Transmission}
This part presents our novel paradigm for feature transmission called Symmetric Uncertainty-aware Feature Transmission (SUFT). Firstly, it is necessary to concatenate RGB features and depth features to transfer the RGB guidance information, which requires the same spatial sizes. Due to the resolution gap, the two features cannot be directly concatenated. A common solution is to upsample the depth map to the same resolution as the corresponding RGB image by bicubic interpolation before they are fed into the network. The drawback of this solution is that interpolation may lead to noise amplification and blurring while enlarging the resolution. As shown in Figure \textcolor{red}{\ref{fig:SUFT}}, SUFT builds an iterative up-and-down sampling pipeline to mitigate these side effects. Specifically, the input depth features are first copied and flipped horizontally in the spatial dimension, which is prepared for calculating Symmetric Uncertainty. Then both depth features are projected into HR domain. These processes can be formulated as:
\vspace{-0.07cm}
\begin{align}
\left\{\begin{aligned}
F^{depth}_{hr} &= (F^{depth}_{lr})\uparrow_{s},\\
F^{flipped}_{hr} &= (HFlip(F^{depth}_{lr}))\uparrow_{s},
\end{aligned}\right.
\vspace{-1cm}
\label{eq:3}
\end{align}
where $F^{depth}_{lr}$ is the features of the depth branch input to SUFT module, $F^{depth}_{hr}$ is the obtained HR depth features, $F^{flipped}_{hr}$ is the flipped HR depth features, $HFlip(\cdot)$ and $(\cdot)\uparrow_{s}$ represent the horizontal flip operation and the up-projection operation with scaling factor $s$, respectively. Note that the two up-projection units are implemented with several convolutional layers. Convolution makes a difference between $F^{depth}_{hr}$ and $F^{flipped}_{hr}$ at the corresponding positions, and the difference is used to calculate the uncertainty. The up-projection units make the spatial size of the two features consistent, and projecting the depth features into HR domain can calculate the uncertainty more accurately. 

Next, depth features are concatenated with RGB features processed by Symmetric Uncertainty:
\vspace{-0.1cm}
\begin{equation} 
\vspace{-0.1cm}
F^{fused}_{hr} = [F^{depth}_{hr};SU(F^{depth}_{hr}, F^{flipped}_{hr}) \cdot F^{RGB}_{hr}],\label{eq:4}
\end{equation}
where $F^{RGB}_{hr}$ is the RGB features extracted by the RGB branch, $F^{fused}_{hr}$ is the fused features, $SU(\cdot)$ denotes the Symmetric Uncertainty, and $[\cdot;\cdot]$ represents the channel-wise concatenation operation. After fusion, the features are mapped back to LR domain by a down-projection unit, which is formulated as:
\vspace{-0.1cm}
\begin{equation}
\vspace{-0.1cm}
F^{fused}_{lr} = (F^{fused}_{hr}) \downarrow_{s},
\end{equation}
where $(\cdot)\downarrow_{s}$ represents the down-projection operation with scaling factor $s$, ensuring that the output of the SUFT module is the same spatial size as the input so that multi-stage feature fusion can be performed. The specific implementation of the up-projection unit and down-projection unit can refer to \cite{CVPR2018DBPN}. The output features $F^{fused}_{lr}$ will be used as input for the next layer in the depth branch. The network can compute the reconstruction error by iterative up-and-down sampling and fine-tune the reconstruction results with multiple feature fusion.

\begin{figure}[tbp]
  \centering
  \includegraphics[width=\linewidth]{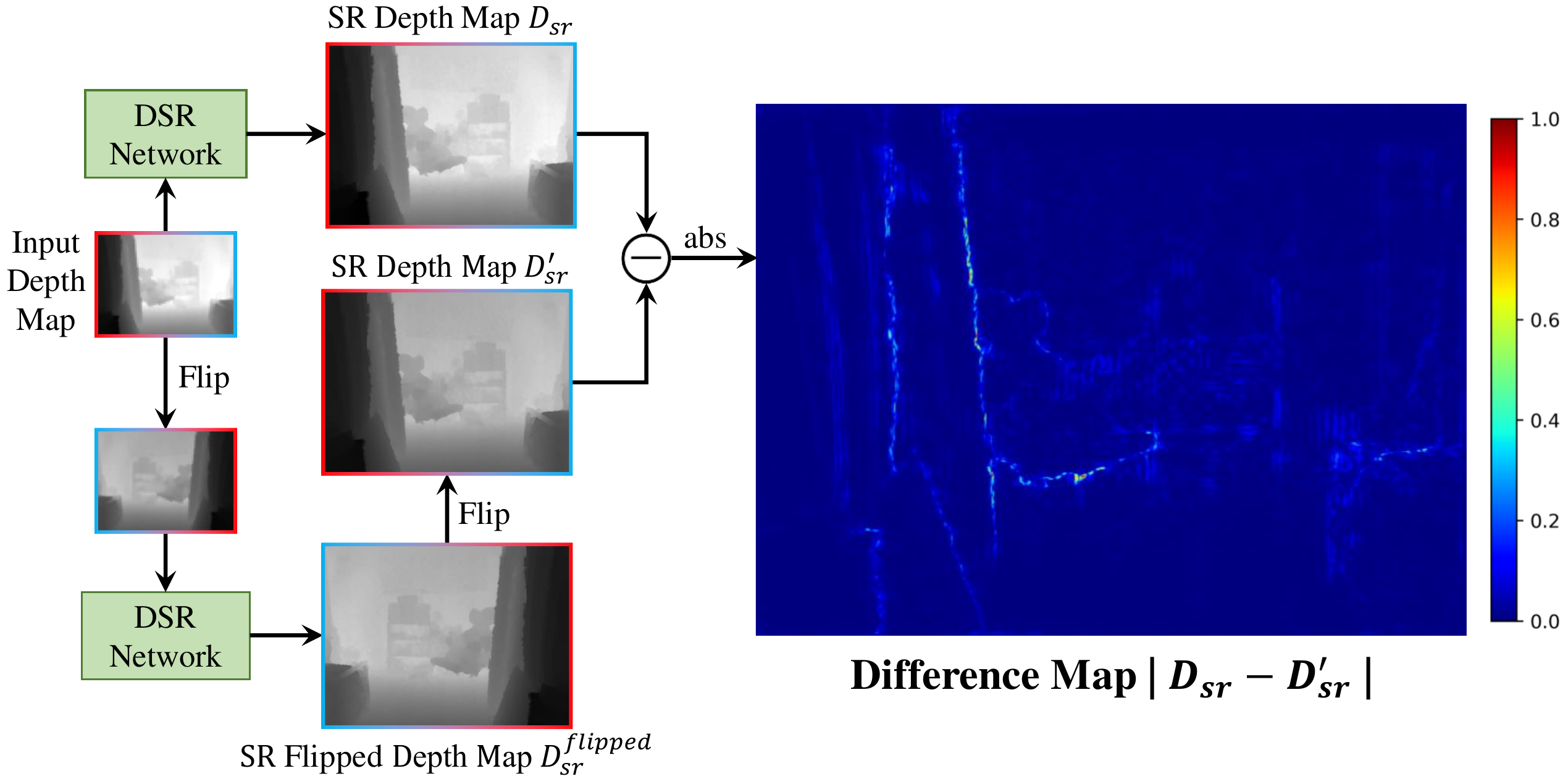}
   \vspace{-0.7cm}
  \caption{Illustration of the rationale of Symmetric Uncertainty and the experiment to compute Symmetric Uncertainty of super-resolved depth map in the pixel space. The difference map shows the absolute difference between \boldmath$D_{sr}$ and $D^{flipped}_{sr}$. Best viewed in color.}
  \label{fig:discussion}
  \vspace{-0.6cm}
\end{figure}

\textbf{Symmetric Uncertainty.} 
Since the cross-modality gap between RGB images and depth maps, it is not appropriate to directly transfer the RGB features or the extracted RGB high-frequency features to the depth branch. For color-guided DSR, color guidance needs to be introduced for edge and texture regions in the depth map rather than in the RGB image. Thus, we propose a novel Symmetric Uncertainty scheme to estimate and remove the RGB features that do not match the depth texture. 
For regions with larger uncertainty in the depth map, the RGB features in the corresponding regions are given higher weights during feature transmission and vice versa to help recover edges and complex textures of depth maps without introducing extra noise. Existing methods for modeling uncertainty cannot generate uncertainty maps and weight the RGB features in a single forward propagation. Our scheme is based on a simple yet effective flipping operation in the feature space, which can be escaped from the above troubles and integrated into the feature transmission. First, we have performed a horizontal flip operation on depth features and projected them into HR domain in Eq.(\textcolor{red}{\ref{eq:3}}). Then we take the absolute difference between the two horizontally mirrored depth features:
\vspace{-0.1cm}
\begin{equation} 
\vspace{-0.1cm}
F^{diff}_{hr} = \left | F^{depth}_{hr} - HFlip(F^{flipped}_{hr}) \right | , 
\end{equation}
where $F^{depth}_{hr}$ and $F^{flipped}_{hr}$ are the input of calculating Symmetric Uncertainty in Eq.(\textcolor{red}{\ref{eq:4}}), and $F^{diff}_{hr}$ represents the absolute difference between $F^{diff}_{hr}$ and $F^{flipped}_{hr}$. By flipping $F^{flipped}_{hr}$ horizontally again, the two features are spatially aligned so that we can calculate the original uncertainty by taking the absolute value after the element-wise subtraction operation.

Then, we aggregate channel information of $F^{diff}_{hr} \in \mathbb{R}^{C \times sH \times sW}$ by applying average-pooling and max-pooling operations along the channel axis, generating two 2D maps: $F^{avg}_{hr} \in \mathbb{R}^{1 \times sH \times sW}$ and $F^{max}_{hr} \in \mathbb{R}^{1 \times sH \times sW}$. Pooling operations along the channel axis have been demonstrated to be helpful in identifying informative locations \cite{ICLR2017attention}. Those are then concatenated and a standard convolution layer convolves them to generate a 2D Symmetric Uncertainty map. The values of Symmetric Uncertainty map are normalized to the range of [0, 1] finally. In short, the process can be formulated as follows:
\vspace{-0.1cm}
\begin{equation}
\begin{split}
\vspace{-0.1cm}
M^{u}_{hr} &= Norm(Conv([AvgPool(F^{diff}_{hr});MaxPool(F^{diff}_{hr})]))\\
&= Norm(Conv([F^{avg}_{hr};F^{max}_{hr}])),
\end{split}
\end{equation}
where $M^{u}_{hr}$ denotes the normalized Symmetric Uncertainty map. The specific operation of $Norm(\cdot)$ can be expressed as:
\vspace{-0.1cm}
\begin{equation}
\vspace{-0.1cm}
X_{norm} = \frac{X - min}{(max - min) + \epsilon }, 
\end{equation}
where $\epsilon$ is a small value to avoid division by zero and defaulting to $1e^{-12}$, $X_{norm}$ is the normalized data, $min$ and $max$ represent the minimum value and maximum value of input data $X$, respectively. After obtaining the Symmetric Uncertainty map $M^{u}_{hr}$, it is used to multiply with the RGB features described in Eq.(\textcolor{red}{\ref{eq:4}}), which will give higher weights to the RGB features corresponding to the regions with larger uncertainty in the depth map and vice versa. As a result, the parts of RGB features that do not match the depth textures are masked out so that texture-copying artifacts can be mitigated.

\begin{figure*}[tbp]
  \centering
  \includegraphics[width=\textwidth]{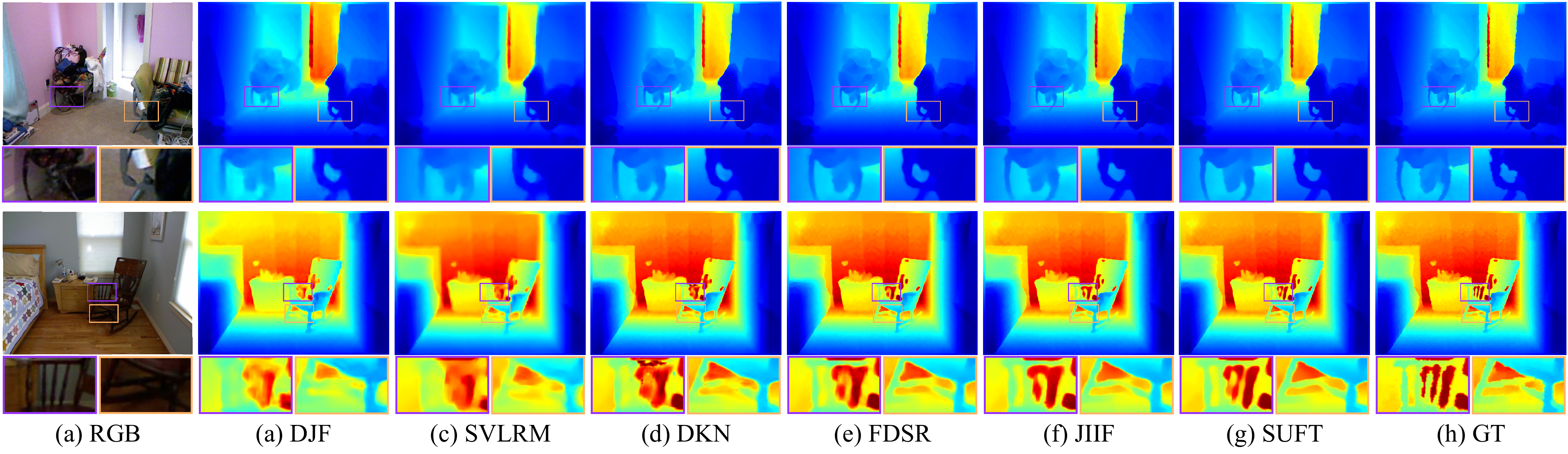}
  \vspace{-0.8cm}
  \caption{Visual comparison of $\times$8 DSR results on NYU v2 dataset. (a) RGB images. (b) DJF \cite{ECCV2016DJF}. (c) SVLRM \cite{CVPR2019SVLRM}. (d) DKN \cite{arxiv2019DKN}. (e) FDSR \cite{CVPR2021FDSR}. (f) JIIF \cite{MM21JIIF}. (g) SUFT. (h) GT. Best viewed in color.}
  \vspace{-0.3cm}
  \label{exp:nyu}
\end{figure*}

\begin{table*}[tbp]
  \caption{Quantitative DSR results (in average RMSE) on NYU v2 dataset. The average RMSE is measured in centimeter. The best performance is displayed in bold, and the second best performance is underlined.}
  \vspace{-0.4cm}
  \label{tab:NYU_v2}
  \setlength\tabcolsep{2.5pt}
  \begin{tabular}{c|ccccccccccccc}
    \toprule
    RMSE & Bicubic & MRF \cite{TIP2015MRF} & TGV \cite{ICCV2013TGV} & SDF \cite{TPAMI2017SDF}  & DG \cite{CVPR2017DGDIE} & SVLRM \cite{CVPR2019SVLRM} & DKN \cite{arxiv2019DKN} & FDSR \cite{CVPR2021FDSR} & CTKT \cite{CVPR2021CTKT} & JIIF \cite{MM21JIIF} & BridgeNet \cite{MM2021bridgenet} & SUFT \\
    \midrule
    $\times$4 & 8.16 & 7.84 & 6.98 & 3.04 & 1.56 & 1.74 & 1.62 & 1.61 & 1.49 & \underline{1.37} & 1.54 & \textbf{1.12} \\
    $\times$8 & 14.22 & 13.98 & 11.23 & 5.67 & 2.99 & 5.59 & 3.26 & 3.18 & 2.73 & 2.76 & \underline{2.63} & \textbf{2.51} \\
    $\times$16 & 22.32 & 22.20 & 28.31 & 9.97 & 5.24 & 7.23 & 6.51 & 5.86 & 5.11 & 5.27 & \underline{4.98} & \textbf{4.86} \\
    \bottomrule
  \end{tabular}
  \vspace{-0.4cm}
\end{table*}

\subsection{Discussions: Why flipping is useful?}
\label{sec:motivation}
Depth super-resolution is an ill-posed problem, as one input LR depth map can be originated from a variety of HR depth maps. For any pixel in the predicted HR image, all possible outcomes for each pixel constitute a distribution $U_{(i, j)}(\mu ,\sigma^2)$, where $(i, j)$ denotes the spatial coordinate of the pixel, $\mu$ and $\sigma^2$ represent the mean and variance of the distribution, respectively. The uncertainty of each pixel is bound up with its variance.

The experiment shown in Figure \textcolor{red}{\ref{fig:discussion}} visually represents our idea of estimating uncertainty. Given an LR depth map and an arbitrary DSR network (the RGB image is omitted here, which has no impact on our discussion), we feed both the LR depth map and the horizontally flipped LR depth map into the network to obtain two horizontally mirrored HR depth maps $D_{sr}$ and $D^{flipped}_{sr}$. After flipping $D^{flipped}_{sr}$ horizontally again, the two depth maps $D_{sr}$ and $D^{'}_{sr}$ are spatially aligned. However, due to the ill-posedness of the super-resolution problem, the depth values of the two reconstructed depth maps are not exactly equal. The super-resolved depth maps $D_{sr}$ and $D^{'}_{sr}$ are equivalent to sampling all possible corresponding HR results twice. The difference between the two samples is generated by convolution, and the spatial transformation increases the difference enough to characterize uncertainty. Our method is not limited to horizontal flip, and the specific way of spatial transformation is not the focus. 
We estimate the spatial distribution of the standard deviation to characterize uncertainty by computing the pixel-wise absolute difference between the two samples:
\begin{align}
    \sigma_{(i, j)} \approx \left | D_{sr}(i, j) - D^{'}_{sr}(i, j)) \right |, 
\end{align}
where $\sigma_{(i, j)}$ is the standard deviation of the distribution of pixel $(i, j)$. The difference map depicts the absolute difference between the two super-resolved depth maps. It can be seen that the uncertainty in the texture and edge regions is much larger than in the smooth regions. This phenomenon is consistent with the results observed in \cite{NuerIPS2021UDL}, proving the validity of our Symmetric Uncertainty. Experiments comparing different spatial transformations can be found in the supplementary material.

\section{Experiments}
\label{Sec 4}
\subsection{Training and Implementation Details}
\label{sec:trainning}
To evaluate the performance of our method, we conduct sufficient experiments on three public datasets, \textit{i.e.}, NYU v2 \cite{ECCV2012NYUv2}, Middlebury and RGB-D-D \cite{CVPR2021FDSR}. We train our SUFT network on NYU v2 dataset and test it on the above three datasets to verify the performance and generalization ability of our method. NYU v2 dataset consists of 1,449 RGB-D pairs. We select the first 1000 RGB-D pairs from NYU v2 dataset for training and the rest 449 RGB-D pairs for testing, following the common splitting method \cite{arxiv2019DKN}. We also use 30 RGB-D pairs from Middlebury 2001 \cite{IJCV2011Middlebury2001}, 2005 \cite{CVPR2007Middlebury2005}, and 2006 \cite{CVPR2007Middlebury2006} datasets provided by Lu \textit{et al.} \cite{CVPR2014Lu} for testing. For RGB-D-D dataset, 405 RGB-D pairs are split as testset following \cite{CVPR2021FDSR}. HR depth maps are cropped into patches of a fixed size of 256 $\times$ 256, which can speed up the training without weakening the performance of the network. In the experiments above, LR depth maps are obtained by bicubic downsampling. Moreover, we evaluate our method in \textit{real-world manner} on RGB-D-D dataset to analyze its effectiveness in real-world scenarios, which contains 2215 RGB-D pairs for training and 405 pairs for testing. In the \textit{real-world manner}, the DSR task is more challenging because the LR depth maps obtained by a low-power Time of Flight (ToF) camera on a mobile phone generally contain some noise even depth holes in addition to downsampling. It makes the mapping between the LR and HR depth map pairs more difficult for the network to learn.

The Root Mean Square Error (RMSE) is employed as the evaluation metric. Lower RMSE values imply better reconstruction quality. We implement our SUFT network with Pytorch and train with an NVIDIA 3090 GPU. During training, the batch size is 1 and the model is optimized using the Adam optimizer with $\beta_1 = 0.9$, $\beta_2 = 0.999$ and $\epsilon = 1e^{-8}$. The initial learning rate is $1e^{-4}$ and reduced by 0.1 for every 100 epochs.

\begin{figure*}[htbp]
  \centering
  \includegraphics[width=\textwidth]{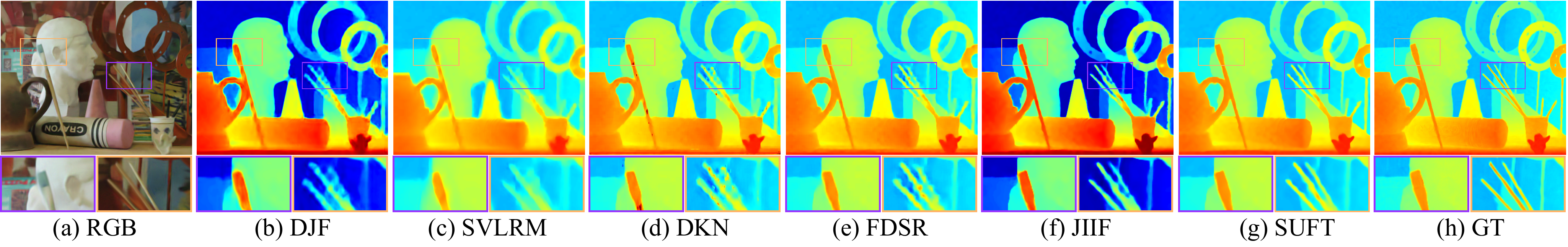}
  \vspace{-0.8cm}
  \caption{Visual comparison of $\times$8 DSR results on Middlebury dataset. (a) RGB images. (b) DJF \cite{ECCV2016DJF}. (c) SVLRM \cite{CVPR2019SVLRM}. (d) DKN \cite{arxiv2019DKN}. (e) FDSR \cite{CVPR2021FDSR}. (f) JIIF \cite{MM21JIIF}. (g) SUFT. (h) GT. Best viewed in color.}
  \vspace{-0.3cm}
  \label{exp:middlebury}
\end{figure*}

\begin{table*}[htbp]
  \caption{Quantitative DSR results (in average RMSE) on Middlebury dataset. The average RMSE is measured in the original scale of the provided disparity. The best performance is displayed in bold, and the second best performance is underlined.}
  \vspace{-0.4cm}
  \label{tab:Middlebury}
  \setlength\tabcolsep{3.3pt}
  \begin{tabular}{c|ccccccccccccc}
    \toprule
    RMSE & Bicubic & DMSG \cite{ECCV2016MSGNet} & DJF \cite{ECCV2016DJF} & DG \cite{CVPR2017DGDIE} & DJFR \cite{TPAMI2019DJFR} & PAC \cite{CVPR2019PAC} & DKN \cite{arxiv2019DKN} & FDKN \cite{arxiv2019DKN} & CUNet \cite{TPAMI2020CUNet} & FDSR \cite{CVPR2021FDSR} & JIIF \cite{MM21JIIF} & SUFT \\
    \midrule
    $\times$4 & 2.28 & 1.88 & 1.68 & 1.97 & 1.32 & 1.32 & 1.23 & \underline{1.08} & 1.10 & 1.13 & 1.09 & \textbf{1.07} \\
    $\times$8 & 3.98 & 3.45 & 3.24 & 4.16 & 3.19 & 2.62 & 2.12 & 2.17 & 2.17 & 2.08 & \underline{1.82} & \textbf{1.75} \\
    $\times$16 & 6.37 & 6.28 & 5.62 & 5.27 & 5.57 & 4.58 & 4.24 & 4.50 & 4.33 & 4.39 & \underline{3.31} & \textbf{3.18} \\
    \bottomrule
  \end{tabular}
  \vspace{-0.2cm}
\end{table*}

\begin{figure*}[htbp]
  \centering
  \includegraphics[width=\textwidth]{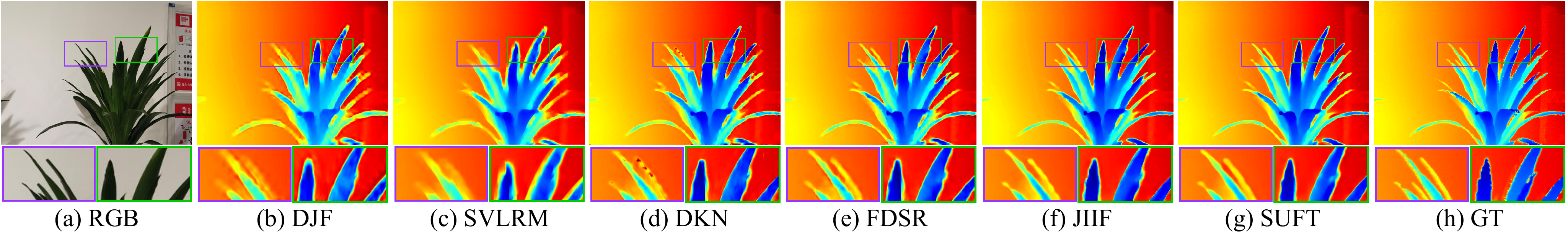}
  \vspace{-0.8cm}
  \caption{Visual comparison of $\times$8 DSR results on RGB-D-D dataset. (a) RGB images. (b) DJF \cite{ECCV2016DJF}. (c) SVLRM \cite{CVPR2019SVLRM}. (d) DKN \cite{arxiv2019DKN}. (e) FDSR \cite{CVPR2021FDSR}. (f) JIIF \cite{MM21JIIF}. (g) SUFT. (h) GT. Best viewed in color.}
  \vspace{-0.3cm}
  \label{exp:rgbdd}
\end{figure*}

\begin{table*}[htbp]
  \caption{Quantitative DSR results (in average RMSE) on RGB-D-D dataset. The average RMSE is measured in centimeter. The best performance is displayed in bold, and the second best performance is underlined.}
  \vspace{-0.4cm}
  \label{tab:RGBDD}
  \setlength\tabcolsep{5pt}
  \begin{tabular}{c|ccccccccccc}
    \toprule
    RMSE & Bicubic & SDF \cite{TPAMI2017SDF} & SVLRM \cite{CVPR2019SVLRM} & DJF \cite{ECCV2016DJF} & DJFR \cite{TPAMI2019DJFR} & PAC \cite{CVPR2019PAC} & DKN \cite{arxiv2019DKN} & FDKN \cite{arxiv2019DKN} & FDSR \cite{CVPR2021FDSR} & JIIF \cite{MM21JIIF} & SUFT \\
    \midrule
    $\times$4 & 2.00 & 4.06 & 3.39 & 3.41 & 3.35 & 1.25 & 1.30 & 1.18 & \underline{1.16} & 1.17 & \textbf{1.10} \\
    $\times$8 & 3.23 & 5.51 & 5.59 & 5.57 & 5.57 & 1.98 & 1.96 & 1.91 & 1.82 & \underline{1.79} & \textbf{1.69} \\
    $\times$16 & 5.16 & 7.39 & 8.28 & 8.15 & 7.99 & 3.49 & 3.42 & 3.41 & 3.06 & \underline{2.87} & \textbf{2.71} \\
    \bottomrule
  \end{tabular}
  \vspace{-0.4cm}
\end{table*}

\vspace{-0.4cm}
\subsection{Performance Comparison}
We compare with some traditional DSR methods as well as some state-of-the-art DSR methods on the three public datasets.

\textbf{NYU v2 Dataset.} For the NYU v2 dataset, we compare our SUFT network with SOTA methods under three different scaling factors ($\times$4, $\times$8, and $\times$16). The qualitative results are reported in Table \textcolor{red}{\ref{tab:NYU_v2}}, where the RMSE is measured in centimeters. With the proposed SUFT, our method achieves the best results on NYU v2 dataset under all scaling factors. Compared with the suboptimal method, the average RMSE of our network reaches 1.12 under the $\times$4 DSR case, with an improvement of 18.2$\%$, and even achieves 4.86 under the most difficult $\times$16 DSR case, with an improvement of 2.4$\%$.

Visual results of our method under the scaling factor of $\times$8 are shown in Figure \textcolor{red}{\ref{exp:nyu}}. In terms of the overall and details of the results, the proposed SUFT can reconstruct the depth information more accurately without introducing texture-copy artifacts and extra noise, which shows that our Symmetric Uncertainty effectively narrows the cross-modality gap.

\textbf{Middlebury Dataset.} We also evaluate our method on Middlebury dataset. Following \cite{arxiv2019DKN}, we train our model on NYU v2 dataset and test it on Middlebury dataset. The quantitative comparisons with recent learning-based methods are summarized in Table \textcolor{red}{\ref{tab:Middlebury}}, where the average RMSE is measured in the original scale of the provided disparity. Our algorithm shows competitive performance under the scaling factor of $\times$4 and goes beyond all comparison methods under the scaling factors of $\times$8 and $\times$16. Figure \textcolor{red}{\ref{exp:middlebury}} presents visual results under the scaling factor of $\times$8. Whether in the purple or green rectangular area, our model generates clear complicated structures and edges that are closest to the ground truth, which demonstrates the superior performance and good generalization ability of our SUFT network.

\textbf{RGB-D-D Dataset.} To demonstrate the effectiveness of SUFT, we further evaluate our method on a novel RGBD dataset: the RGB-D-D dataset. We train our network on NYU v2 dataset and test it on RGB-D-D dataset, which contains 405 RGB-D pairs for testing. Table \textcolor{red}{\ref{tab:RGBDD}} shows the quantitative results in terms of average RMSE measured in centimeters. To more intuitively illustrate the advantages of our method, we provide a visual comparison of the $\times$8 DSR results in Figure \textcolor{red}{\ref{exp:rgbdd}}.

\begin{table*}[tbp]
  \caption{Quantitative DSR results (in average RMSE) in \textit{real-world manner} on RGB-D-D dataset. The average RMSE is measured in centimeter. FDSR$^{\ast}$, JIIF$^{\ast}$, and SUFT$^{\ast}$ are trained in \textit{real-world manner}.}
  \vspace{-0.3cm}
  \label{tab:real_world}
  \setlength\tabcolsep{3.3pt}
  \begin{tabular}{c|cccccccccccc}
    \toprule
    RMSE & Bicubic & SVLRM \cite{CVPR2019SVLRM} & DJF \cite{ECCV2016DJF} & DJFR \cite{TPAMI2019DJFR} & FDKN \cite{arxiv2019DKN} & DKN \cite{arxiv2019DKN} & FDSR \cite{CVPR2021FDSR} & JIIF \cite{MM21JIIF} & SUFT & FDSR$^{\ast}$ \cite{CVPR2021FDSR} & JIIF$^{\ast}$ \cite{MM21JIIF} & SUFT$^{\ast}$ \\
    \midrule
    \textit{real} & 9.15 & 8.50 & 7.90 & 8.01 & 7.50 & 7.38 & 7.50 & 8.41 & 7.22 & \underline{5.49} & 5.83 & \textbf{5.41}\\
    \bottomrule
  \end{tabular}
  \vspace{-0.3cm}
\end{table*}

\begin{figure}[tbp]
  \centering
  \includegraphics[width=\linewidth]{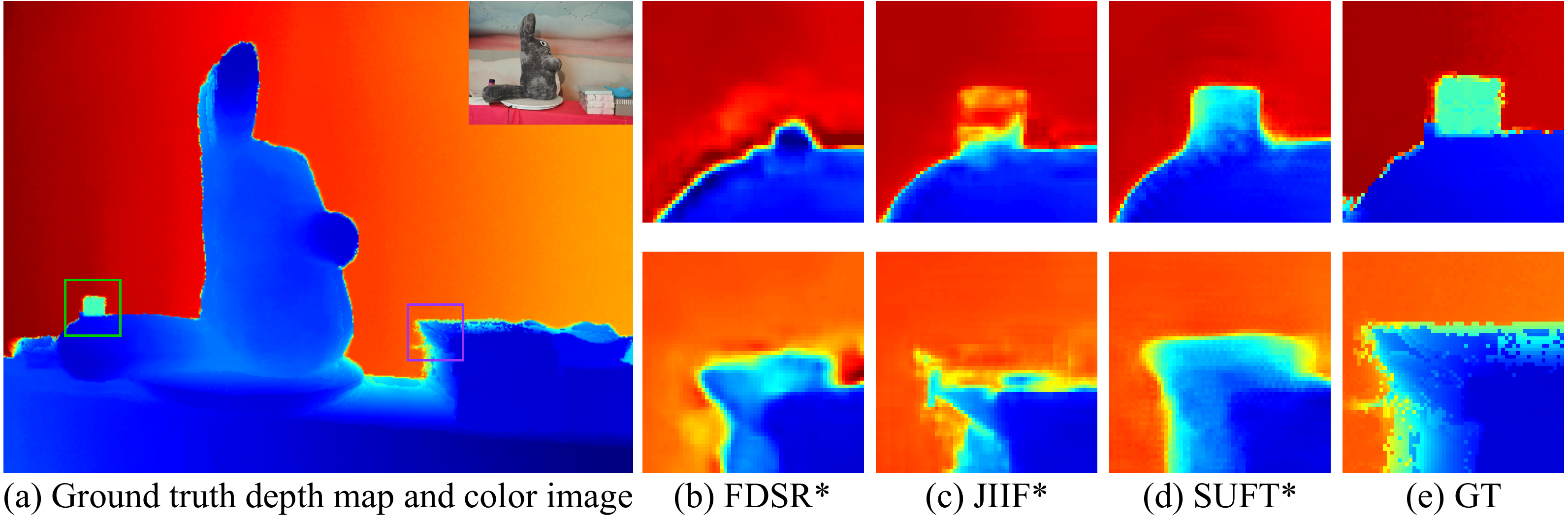}
  \vspace{-0.7cm}
  \caption{Visual comparison of $\times$8 DSR results in the \textit{real-world manner} on RGB-D-D dataset. (a) Ground truth depth map and color image. (b) FDSR$^{\ast}$ \cite{CVPR2021FDSR}. (c) JIIF$^{\ast}$ \cite{MM21JIIF}. (d) SUFT$^{\ast}$. (e) GT. Best viewed in color.}
  \label{exp:rgbdd_real}
  \vspace{-0.3cm}
\end{figure}

Moreover, to explore the performance of our method in real scenarios, we experimented in \textit{real-world manner} on RGB-D-D dataset. The size of the LR depth map and the target HR depth map are 192 $\times$ 144 and 512 $\times$ 384, respectively. Following \cite{CVPR2021FDSR}, the existing $\times$4 models in Table \textcolor{red}{\ref{tab:RGBDD}} are used for evaluation. Likewise, we also provide a version of SUFT network trained in \textit{real-world manner} for a fair comparison. When training in \textit{real-world manner}, the initial learning rate is $6e^{-5}$ and reduced by 0.5 for every 70 epochs and we scale the depth maps to the closest scaling factor of $\times$2 before feeding the RGB-D pairs to the network. The results are appended in Table \textcolor{red}{\ref{tab:real_world}} and Figure \textcolor{red}{\ref{exp:rgbdd_real}}, where we only show the visual results of the method trained in \textit{real-world manner}. 

As shown in Table \textcolor{red}{\ref{tab:RGBDD}}, our SUFT network achieves the best performance under all upsampling cases. Even at the most difficult scaling factor of $\times$16, our method refreshes the average RMSE from 2.87 to 2.71 and improves by 5.6\% compared to recent SOTA work. In the more challenging \textit{real-world manner}, our model also achieves the best results in both settings. All experiments on RGB-D-D dataset demonstrate the robustness of our method and its potential to handle practical DSR tasks in real-world scenarios.

\begin{table}[tbp]
  \caption{Ablation studies (in average RMSE) of our SUFT on the NYU v2 dataset and RGB-D-D dataset ($\times$8 case).}
  \vspace{-0.3cm}
  \label{tab:ablation}
  \centering
    \begin{tabular}{ccc|c|c}
    \toprule
\multirow{2}{*}{\begin{tabular}[c]{@{}c@{}}Baseline\\ Model\end{tabular}} & \multirow{2}{*}{\begin{tabular}[c]{@{}c@{}}Iterative\\ Upsampling\end{tabular}} & \multirow{2}{*}{\begin{tabular}[c]{@{}c@{}}Symmetric\\ Uncertainty\end{tabular}} & \multirow{2}{*}{NYU v2} & \multirow{2}{*}{RGB-D-D} \\
                    &                      &                         &                      &                       \\ \midrule
    $\checkmark$    &                      &                         &    2.79              &    1.87               \\
    $\checkmark$    &  $\checkmark$        &                         &    2.59              &    1.75               \\
    $\checkmark$    &                      &  $\checkmark$           &    2.64              &    1.81               \\ \midrule
    $\checkmark$    &  $\checkmark$        &  $\checkmark$           &    \textbf{2.51}     &    \textbf{1.69}      \\
    \bottomrule
\end{tabular}
\vspace{-0.4cm}
\end{table}

\subsection{Ablation Study}
In this section, we conduct several ablation studies to analyze the impact of the core designs in our method. We compare the $\times$8 DSR results on NYU v2 dataset and RGB-D-D dataset under different experimental settings in Table \textcolor{red}{\ref{tab:ablation}}. To illustrate the effectiveness of SUFT, in the baseline model we replace iterative up-and-down sampling with interpolated pre-upsampling and transmit the features extracted from the RGB branch to the depth branch without any processing. Next, we apply iterative up-and-down sampling to the baseline model. As shown in the $1^{st}$ and $2^{nd}$ rows of Table \textcolor{red}{\ref{tab:ablation}}, the DSR result on NYU v2 dataset is improved from 2.79 to 2.59. Then, we integrate Symmetric Uncertainty into the framework to assess its contribution. When the baseline model uses only Symmetric Uncertainty, the result is improved to 2.64. Note that in this case we have replaced the up-projection unit and down-projection unit with several ‘Conv+ReLU’ layers for a fair comparison. Furthermore, when the two designs are used in combination (\textit{i.e.}, full SUFT module) with the baseline model, the best performance is achieved. The experimental results on RGB-D-D datasets are similar.

\begin{figure}[tbp]
  \centering
  \includegraphics[width=\linewidth]{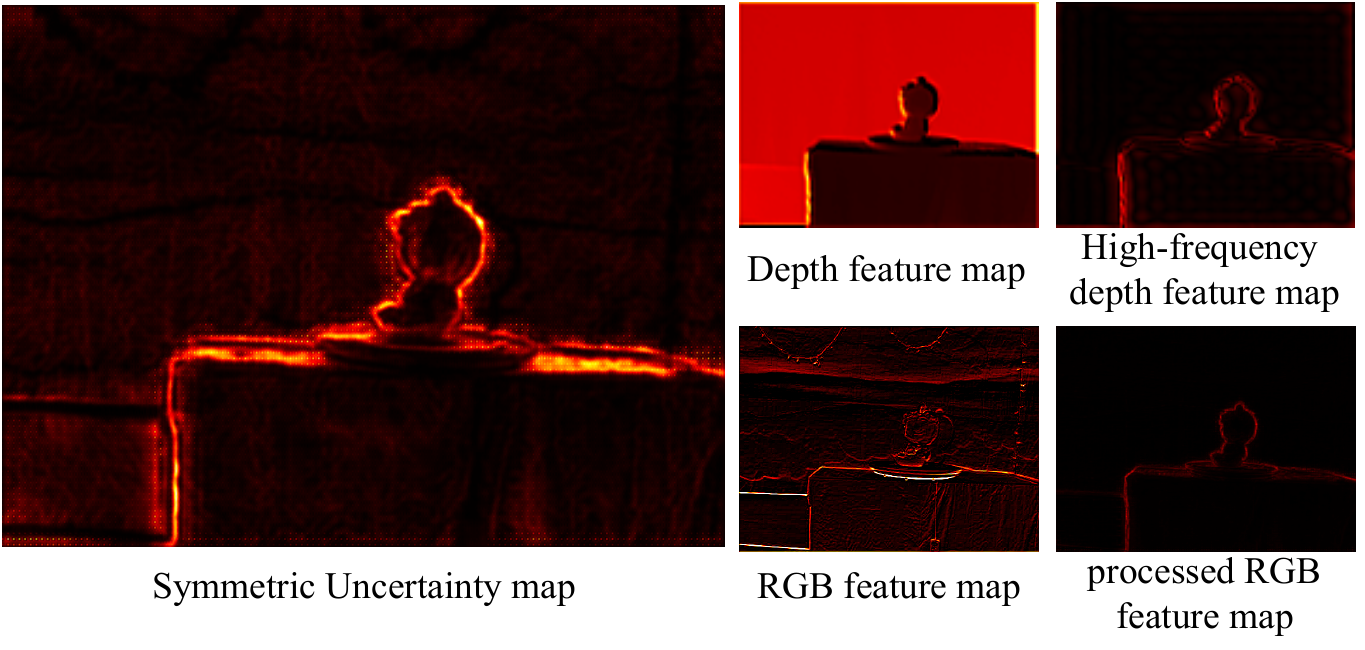}
  \vspace{-0.7cm}
  \caption{Visualization of ablation study on Symmetric Uncertainty. There is a significant difference between the original RGB features and depth features. However, RGB features processed by Symmetric Uncertainty are very close to the high-frequency components of the depth features and have sharp edges. Best viewed in color.}
  \label{fig:ablation}
  \vspace{-0.4cm}
\end{figure}

To further analyze the effect of Symmetric Uncertainty, we conduct visualization experiments shown in Figure \textcolor{red}{\ref{fig:ablation}}. The first column is the Symmetric Uncertainty map, which highlights the hard-to-recover areas in the depth map. The cross-modality gap can be clearly seen by comparing the depth feature map and the original RGB feature map (not processed by Symmetric Uncertainty). We use a high-pass filter to obtain the high-frequency components of the depth feature map. It is noticeable that the original RGB features contain some high-frequency information that does not belong to the depth modality. Besides, we also visualize the RGB feature map processed by Symmetric Uncertainty, which is very close to the high-frequency components of the depth feature map. Compared with simply transmitting unprocessed RGB features to the depth branch, our proposed Symmetric Uncertainty can provide more effective guidance and avoid triggering texture-copying artifacts.

\section{Conclusion}
\label{Sec 5}
In this paper, we propose a novel Symmetric Uncertainty-aware Feature Transmission (SUFT) that bridges the resolution gap and the cross-modality gap to boost the color-guided DSR performance. On the one hand, we build an iterative up-and-down sampling pipeline instead of interpolated pre-upsampling to bridge the resolution gap, which alleviates noise amplification and blurring. On the other hand, we propose a novel Symmetric Uncertainty scheme to process the extracted RGB features to narrow the cross-modality gap in reducing the texture-copying artifacts in the reconstruction results. Further inserting SUFT into a multi-stage fusion network, our method creates new state-of-the-arts on multiple datasets. Extensive evaluations demonstrate that SUFT is superior to previous methods and facilitates accurate depth super-resolution.

\begin{acks}
This work is supported by National Natural Science Foundation of China (62176188), Key Research and Development Program of Hubei Province (2021BAA187), Special Fund of Hubei Luojia Laboratory (220100015).
\end{acks}


\clearpage
\bibliographystyle{ACM-Reference-Format}
\bibliography{reference}


\begin{thebibliography}{68}


\ifx \showCODEN    \undefined \def \showCODEN     #1{\unskip}     \fi
\ifx \showDOI      \undefined \def \showDOI       #1{#1}\fi
\ifx \showISBNx    \undefined \def \showISBNx     #1{\unskip}     \fi
\ifx \showISBNxiii \undefined \def \showISBNxiii  #1{\unskip}     \fi
\ifx \showISSN     \undefined \def \showISSN      #1{\unskip}     \fi
\ifx \showLCCN     \undefined \def \showLCCN      #1{\unskip}     \fi
\ifx \shownote     \undefined \def \shownote      #1{#1}          \fi
\ifx \showarticletitle \undefined \def \showarticletitle #1{#1}   \fi
\ifx \showURL      \undefined \def \showURL       {\relax}        \fi
\providecommand\bibfield[2]{#2}
\providecommand\bibinfo[2]{#2}
\providecommand\natexlab[1]{#1}
\providecommand\showeprint[2][]{arXiv:#2}

\bibitem[AlBahar and Huang(2019)]%
        {ICCV2019GbFT}
\bibfield{author}{\bibinfo{person}{Badour AlBahar} {and}
  \bibinfo{person}{Jia-Bin Huang}.} \bibinfo{year}{2019}\natexlab{}.
\newblock \showarticletitle{Guided image-to-image translation with
  bi-directional feature transformation}. In \bibinfo{booktitle}{\emph{ICCV}}.
  \bibinfo{pages}{9016--9025}.
\newblock


\bibitem[Badrinarayanan et~al\mbox{.}(2017)]%
        {TPAMI2017Segnet}
\bibfield{author}{\bibinfo{person}{Vijay Badrinarayanan}, \bibinfo{person}{Alex
  Kendall}, {and} \bibinfo{person}{Roberto Cipolla}.}
  \bibinfo{year}{2017}\natexlab{}.
\newblock \showarticletitle{Segnet: A deep convolutional encoder-decoder
  architecture for image segmentation}.
\newblock \bibinfo{journal}{\emph{IEEE TPAMI}} (\bibinfo{year}{2017}),
  \bibinfo{pages}{2481--2495}.
\newblock


\bibitem[Baker et~al\mbox{.}(2011)]%
        {IJCV2011Middlebury2001}
\bibfield{author}{\bibinfo{person}{Simon Baker}, \bibinfo{person}{Daniel
  Scharstein}, \bibinfo{person}{JP Lewis}, \bibinfo{person}{Stefan Roth},
  \bibinfo{person}{Michael~J Black}, {and} \bibinfo{person}{Richard Szeliski}.}
  \bibinfo{year}{2011}\natexlab{}.
\newblock \showarticletitle{A database and evaluation methodology for optical
  flow}.
\newblock \bibinfo{journal}{\emph{IJCV}} (\bibinfo{year}{2011}),
  \bibinfo{pages}{1--31}.
\newblock


\bibitem[Bishop and Quazaz(1996)]%
        {NuerIPS1996BNN}
\bibfield{author}{\bibinfo{person}{Christopher Bishop} {and}
  \bibinfo{person}{Cazhaow Quazaz}.} \bibinfo{year}{1996}\natexlab{}.
\newblock \showarticletitle{Regression with input-dependent noise: A Bayesian
  treatment}.
\newblock \bibinfo{journal}{\emph{NuerIPS}}.
\newblock


\bibitem[Chen et~al\mbox{.}(2019)]%
        {ICCV2019OCConv}
\bibfield{author}{\bibinfo{person}{Yunpeng Chen}, \bibinfo{person}{Haoqi Fan},
  \bibinfo{person}{Bing Xu}, \bibinfo{person}{Zhicheng Yan},
  \bibinfo{person}{Yannis Kalantidis}, \bibinfo{person}{Marcus Rohrbach},
  \bibinfo{person}{Shuicheng Yan}, {and} \bibinfo{person}{Jiashi Feng}.}
  \bibinfo{year}{2019}\natexlab{}.
\newblock \showarticletitle{Drop an octave: Reducing spatial redundancy in
  convolutional neural networks with octave convolution}. In
  \bibinfo{booktitle}{\emph{ICCV}}. \bibinfo{pages}{3435--3444}.
\newblock


\bibitem[Cong et~al\mbox{.}(2017a)]%
        {TIP2017co}
\bibfield{author}{\bibinfo{person}{Runmin Cong}, \bibinfo{person}{Jianjun Lei},
  \bibinfo{person}{Huazhu Fu}, \bibinfo{person}{Qingming Huang},
  \bibinfo{person}{Xiaochun Cao}, {and} \bibinfo{person}{Chunping Hou}.}
  \bibinfo{year}{2017}\natexlab{a}.
\newblock \showarticletitle{Co-saliency detection for RGBD images based on
  multi-constraint feature matching and cross label propagation}.
\newblock \bibinfo{journal}{\emph{IEEE TIP}} (\bibinfo{year}{2017}),
  \bibinfo{pages}{568--579}.
\newblock


\bibitem[Cong et~al\mbox{.}(2017b)]%
        {TCYB2017iterative}
\bibfield{author}{\bibinfo{person}{Runmin Cong}, \bibinfo{person}{Jianjun Lei},
  \bibinfo{person}{Huazhu Fu}, \bibinfo{person}{Weisi Lin},
  \bibinfo{person}{Qingming Huang}, \bibinfo{person}{Xiaochun Cao}, {and}
  \bibinfo{person}{Chunping Hou}.} \bibinfo{year}{2017}\natexlab{b}.
\newblock \showarticletitle{An iterative co-saliency framework for RGBD
  images}.
\newblock \bibinfo{journal}{\emph{IEEE TCYB}} (\bibinfo{year}{2017}),
  \bibinfo{pages}{233--246}.
\newblock


\bibitem[Cong et~al\mbox{.}(2016)]%
        {SPL2016saliency}
\bibfield{author}{\bibinfo{person}{Runmin Cong}, \bibinfo{person}{Jianjun Lei},
  \bibinfo{person}{Changqing Zhang}, \bibinfo{person}{Qingming Huang},
  \bibinfo{person}{Xiaochun Cao}, {and} \bibinfo{person}{Chunping Hou}.}
  \bibinfo{year}{2016}\natexlab{}.
\newblock \showarticletitle{Saliency detection for stereoscopic images based on
  depth confidence analysis and multiple cues fusion}.
\newblock \bibinfo{journal}{\emph{IEEE SPL}} (\bibinfo{year}{2016}),
  \bibinfo{pages}{819--823}.
\newblock


\bibitem[Deng and Dragotti(2020)]%
        {TPAMI2020CUNet}
\bibfield{author}{\bibinfo{person}{Xin Deng} {and} \bibinfo{person}{Pier~Luigi
  Dragotti}.} \bibinfo{year}{2020}\natexlab{}.
\newblock \showarticletitle{Deep convolutional neural network for multi-modal
  image restoration and fusion}.
\newblock \bibinfo{journal}{\emph{IEEE TPAMI}} (\bibinfo{year}{2020}),
  \bibinfo{pages}{3333--3348}.
\newblock


\bibitem[Ferstl et~al\mbox{.}(2013)]%
        {ICCV2013TGV}
\bibfield{author}{\bibinfo{person}{David Ferstl}, \bibinfo{person}{Christian
  Reinbacher}, \bibinfo{person}{Rene Ranftl}, \bibinfo{person}{Matthias
  R{\"u}ther}, {and} \bibinfo{person}{Horst Bischof}.}
  \bibinfo{year}{2013}\natexlab{}.
\newblock \showarticletitle{Image guided depth upsampling using anisotropic
  total generalized variation}. In \bibinfo{booktitle}{\emph{ICCV}}.
  \bibinfo{pages}{993--1000}.
\newblock


\bibitem[Goldberg et~al\mbox{.}(1997)]%
        {NuerIPS1997Uncertainty}
\bibfield{author}{\bibinfo{person}{Paul Goldberg}, \bibinfo{person}{Christopher
  Williams}, {and} \bibinfo{person}{Christopher Bishop}.}
  \bibinfo{year}{1997}\natexlab{}.
\newblock \showarticletitle{Regression with input-dependent noise: A Gaussian
  process treatment}.
\newblock \bibinfo{journal}{\emph{NuerIPS}}.
\newblock


\bibitem[Gu et~al\mbox{.}(2017)]%
        {CVPR2017DGDIE}
\bibfield{author}{\bibinfo{person}{Shuhang Gu}, \bibinfo{person}{Wangmeng Zuo},
  \bibinfo{person}{Shi Guo}, \bibinfo{person}{Yunjin Chen},
  \bibinfo{person}{Chongyu Chen}, {and} \bibinfo{person}{Lei Zhang}.}
  \bibinfo{year}{2017}\natexlab{}.
\newblock \showarticletitle{Learning dynamic guidance for depth image
  enhancement}. In \bibinfo{booktitle}{\emph{CVPR}}.
  \bibinfo{pages}{3769--3778}.
\newblock


\bibitem[Gu et~al\mbox{.}(2015)]%
        {CVI2015}
\bibfield{author}{\bibinfo{person}{Yingjie Gu}, \bibinfo{person}{Zhong Jin},
  {and} \bibinfo{person}{Steve~C Chiu}.} \bibinfo{year}{2015}\natexlab{}.
\newblock \showarticletitle{Active learning combining uncertainty and diversity
  for multi-class image classification}.
\newblock \bibinfo{journal}{\emph{IET CVI}} (\bibinfo{year}{2015}),
  \bibinfo{pages}{400--407}.
\newblock


\bibitem[Guo et~al\mbox{.}(2018)]%
        {TIP2018DepthSRNet}
\bibfield{author}{\bibinfo{person}{Chunle Guo}, \bibinfo{person}{Chongyi Li},
  \bibinfo{person}{Jichang Guo}, \bibinfo{person}{Runmin Cong},
  \bibinfo{person}{Huazhu Fu}, {and} \bibinfo{person}{Ping Han}.}
  \bibinfo{year}{2018}\natexlab{}.
\newblock \showarticletitle{Hierarchical features driven residual learning for
  depth map super-resolution}.
\newblock \bibinfo{journal}{\emph{IEEE TIP}} (\bibinfo{year}{2018}),
  \bibinfo{pages}{2545--2557}.
\newblock


\bibitem[Ham et~al\mbox{.}(2017)]%
        {TPAMI2017SDF}
\bibfield{author}{\bibinfo{person}{Bumsub Ham}, \bibinfo{person}{Minsu Cho},
  {and} \bibinfo{person}{Jean Ponce}.} \bibinfo{year}{2017}\natexlab{}.
\newblock \showarticletitle{Robust guided image filtering using nonconvex
  potentials}.
\newblock \bibinfo{journal}{\emph{IEEE TMAPI}} (\bibinfo{year}{2017}),
  \bibinfo{pages}{192--207}.
\newblock


\bibitem[Haris et~al\mbox{.}(2018)]%
        {CVPR2018DBPN}
\bibfield{author}{\bibinfo{person}{Muhammad Haris}, \bibinfo{person}{Gregory
  Shakhnarovich}, {and} \bibinfo{person}{Norimichi Ukita}.}
  \bibinfo{year}{2018}\natexlab{}.
\newblock \showarticletitle{Deep back-projection networks for
  super-resolution}. In \bibinfo{booktitle}{\emph{CVPR}}.
  \bibinfo{pages}{1664--1673}.
\newblock


\bibitem[Haris et~al\mbox{.}(2019)]%
        {CVPR2019RBPN}
\bibfield{author}{\bibinfo{person}{Muhammad Haris}, \bibinfo{person}{Gregory
  Shakhnarovich}, {and} \bibinfo{person}{Norimichi Ukita}.}
  \bibinfo{year}{2019}\natexlab{}.
\newblock \showarticletitle{Recurrent back-projection network for video
  super-resolution}. In \bibinfo{booktitle}{\emph{CVPR}}.
  \bibinfo{pages}{3897--3906}.
\newblock


\bibitem[He et~al\mbox{.}(2012)]%
        {TPAMI2012filter2}
\bibfield{author}{\bibinfo{person}{Kaiming He}, \bibinfo{person}{Jian Sun},
  {and} \bibinfo{person}{Xiaoou Tang}.} \bibinfo{year}{2012}\natexlab{}.
\newblock \showarticletitle{Guided image filtering}.
\newblock \bibinfo{journal}{\emph{IEEE TPAMI}} (\bibinfo{year}{2012}),
  \bibinfo{pages}{1397--1409}.
\newblock


\bibitem[He et~al\mbox{.}(2021)]%
        {CVPR2021FDSR}
\bibfield{author}{\bibinfo{person}{Lingzhi He}, \bibinfo{person}{Hongguang
  Zhu}, \bibinfo{person}{Feng Li}, \bibinfo{person}{Huihui Bai},
  \bibinfo{person}{Runmin Cong}, \bibinfo{person}{Chunjie Zhang},
  \bibinfo{person}{Chunyu Lin}, \bibinfo{person}{Meiqin Liu}, {and}
  \bibinfo{person}{Yao Zhao}.} \bibinfo{year}{2021}\natexlab{}.
\newblock \showarticletitle{Towards fast and accurate real-world depth
  super-resolution: Benchmark dataset and baseline}. In
  \bibinfo{booktitle}{\emph{CVPR}}. \bibinfo{pages}{9229--9238}.
\newblock


\bibitem[Hirschmuller and Scharstein(2007)]%
        {CVPR2007Middlebury2006}
\bibfield{author}{\bibinfo{person}{Heiko Hirschmuller} {and}
  \bibinfo{person}{Daniel Scharstein}.} \bibinfo{year}{2007}\natexlab{}.
\newblock \showarticletitle{Evaluation of cost functions for stereo matching}.
  In \bibinfo{booktitle}{\emph{CVPR}}. \bibinfo{pages}{1--8}.
\newblock


\bibitem[Hoffman et~al\mbox{.}(2016)]%
        {CVPR2016scenerecognition}
\bibfield{author}{\bibinfo{person}{Judy Hoffman}, \bibinfo{person}{Saurabh
  Gupta}, {and} \bibinfo{person}{Trevor Darrell}.}
  \bibinfo{year}{2016}\natexlab{}.
\newblock \showarticletitle{Learning with side information through modality
  hallucination}. In \bibinfo{booktitle}{\emph{CVPR}}.
  \bibinfo{pages}{826--834}.
\newblock


\bibitem[Hui et~al\mbox{.}(2016)]%
        {ECCV2016MSGNet}
\bibfield{author}{\bibinfo{person}{Tak-Wai Hui}, \bibinfo{person}{Chen~Change
  Loy}, {and} \bibinfo{person}{Xiaoou Tang}.} \bibinfo{year}{2016}\natexlab{}.
\newblock \showarticletitle{Depth map super-resolution by deep multi-scale
  guidance}. In \bibinfo{booktitle}{\emph{ECCV}}. \bibinfo{pages}{353--369}.
\newblock


\bibitem[Im et~al\mbox{.}(2018)]%
        {TPAMI20183Dreconstruction}
\bibfield{author}{\bibinfo{person}{Sunghoon Im}, \bibinfo{person}{Hyowon Ha},
  \bibinfo{person}{Gyeongmin Choe}, \bibinfo{person}{Hae-Gon Jeon},
  \bibinfo{person}{Kyungdon Joo}, {and} \bibinfo{person}{In~So Kweon}.}
  \bibinfo{year}{2018}\natexlab{}.
\newblock \showarticletitle{Accurate 3d reconstruction from small motion clip
  for rolling shutter cameras}.
\newblock \bibinfo{journal}{\emph{IEEE TPAMI}} (\bibinfo{year}{2018}),
  \bibinfo{pages}{775--787}.
\newblock


\bibitem[Ioffe and Szegedy(2015)]%
        {ICML2015BN}
\bibfield{author}{\bibinfo{person}{Sergey Ioffe} {and}
  \bibinfo{person}{Christian Szegedy}.} \bibinfo{year}{2015}\natexlab{}.
\newblock \showarticletitle{Batch normalization: Accelerating deep network
  training by reducing internal covariate shift}. In
  \bibinfo{booktitle}{\emph{ICML}}. \bibinfo{pages}{448--456}.
\newblock


\bibitem[Kar and Biswas(2021)]%
        {CVPR2021FastBNN}
\bibfield{author}{\bibinfo{person}{Aupendu Kar} {and}
  \bibinfo{person}{Prabir~Kumar Biswas}.} \bibinfo{year}{2021}\natexlab{}.
\newblock \showarticletitle{Fast bayesian uncertainty estimation and reduction
  of batch normalized single image super-resolution network}. In
  \bibinfo{booktitle}{\emph{CVPR}}. \bibinfo{pages}{4957--4966}.
\newblock


\bibitem[Kendall and Gal(2017)]%
        {NuerIPS2017Uncertainty}
\bibfield{author}{\bibinfo{person}{Alex Kendall} {and} \bibinfo{person}{Yarin
  Gal}.} \bibinfo{year}{2017}\natexlab{}.
\newblock \showarticletitle{What uncertainties do we need in bayesian deep
  learning for computer vision?}
\newblock \bibinfo{journal}{\emph{NuerIPS}}.
\newblock


\bibitem[Kerl et~al\mbox{.}(2013)]%
        {IROS2013SLAM}
\bibfield{author}{\bibinfo{person}{Christian Kerl}, \bibinfo{person}{J{\"u}rgen
  Sturm}, {and} \bibinfo{person}{Daniel Cremers}.}
  \bibinfo{year}{2013}\natexlab{}.
\newblock \showarticletitle{Dense visual SLAM for RGB-D cameras}. In
  \bibinfo{booktitle}{\emph{IROS}}. \bibinfo{pages}{2100--2106}.
\newblock


\bibitem[Kim et~al\mbox{.}(2019)]%
        {arxiv2019DKN}
\bibfield{author}{\bibinfo{person}{Beomjun Kim}, \bibinfo{person}{Jean Ponce},
  {and} \bibinfo{person}{Bumsub Ham}.} \bibinfo{year}{2019}\natexlab{}.
\newblock \showarticletitle{Deformable kernel networks for guided depth map
  upsampling}.
\newblock \bibinfo{journal}{\emph{arXiv preprint arXiv:1903.11286}}
  (\bibinfo{year}{2019}).
\newblock


\bibitem[Lee and Chung(2019)]%
        {ICMLA2019GRAM}
\bibfield{author}{\bibinfo{person}{Changwoo Lee} {and} \bibinfo{person}{Ki-Seok
  Chung}.} \bibinfo{year}{2019}\natexlab{}.
\newblock \showarticletitle{Gram: Gradient rescaling attention model for data
  uncertainty estimation in single image super resolution}. In
  \bibinfo{booktitle}{\emph{IEEE ICMLA}}. \bibinfo{pages}{8--13}.
\newblock


\bibitem[Li et~al\mbox{.}(2020a)]%
        {TCYB2020textcopy}
\bibfield{author}{\bibinfo{person}{Chongyi Li}, \bibinfo{person}{Runmin Cong},
  \bibinfo{person}{Sam Kwong}, \bibinfo{person}{Junhui Hou},
  \bibinfo{person}{Huazhu Fu}, \bibinfo{person}{Guopu Zhu},
  \bibinfo{person}{Dingwen Zhang}, {and} \bibinfo{person}{Qingming Huang}.}
  \bibinfo{year}{2020}\natexlab{a}.
\newblock \showarticletitle{ASIF-Net: Attention steered interweave fusion
  network for RGB-D salient object detection}.
\newblock \bibinfo{journal}{\emph{IEEE TCYB}} (\bibinfo{year}{2020}),
  \bibinfo{pages}{88--100}.
\newblock


\bibitem[Li et~al\mbox{.}(2020b)]%
        {ECCV2020depthbleed}
\bibfield{author}{\bibinfo{person}{Chongyi Li}, \bibinfo{person}{Runmin Cong},
  \bibinfo{person}{Yongri Piao}, \bibinfo{person}{Qianqian Xu}, {and}
  \bibinfo{person}{Chen~Change Loy}.} \bibinfo{year}{2020}\natexlab{b}.
\newblock \showarticletitle{RGB-D salient object detection with cross-modality
  modulation and selection}. In \bibinfo{booktitle}{\emph{ECCV}}.
  \bibinfo{pages}{225--241}.
\newblock


\bibitem[Li et~al\mbox{.}(2016)]%
        {ECCV2016DJF}
\bibfield{author}{\bibinfo{person}{Yijun Li}, \bibinfo{person}{Jia-Bin Huang},
  \bibinfo{person}{Narendra Ahuja}, {and} \bibinfo{person}{Ming-Hsuan Yang}.}
  \bibinfo{year}{2016}\natexlab{}.
\newblock \showarticletitle{Deep joint image filtering}. In
  \bibinfo{booktitle}{\emph{ECCV}}. \bibinfo{pages}{154--169}.
\newblock


\bibitem[Li et~al\mbox{.}(2019a)]%
        {TPAMI2019DJFR}
\bibfield{author}{\bibinfo{person}{Yijun Li}, \bibinfo{person}{Jia-Bin Huang},
  \bibinfo{person}{Narendra Ahuja}, {and} \bibinfo{person}{Ming-Hsuan Yang}.}
  \bibinfo{year}{2019}\natexlab{a}.
\newblock \showarticletitle{Joint image filtering with deep convolutional
  networks}.
\newblock \bibinfo{journal}{\emph{IEEE TPAMI}} (\bibinfo{year}{2019}),
  \bibinfo{pages}{1909--1923}.
\newblock


\bibitem[Li et~al\mbox{.}(2019b)]%
        {CVPR2019SRFBN}
\bibfield{author}{\bibinfo{person}{Zhen Li}, \bibinfo{person}{Jinglei Yang},
  \bibinfo{person}{Zheng Liu}, \bibinfo{person}{Xiaomin Yang},
  \bibinfo{person}{Gwanggil Jeon}, {and} \bibinfo{person}{Wei Wu}.}
  \bibinfo{year}{2019}\natexlab{b}.
\newblock \showarticletitle{Feedback network for image super-resolution}. In
  \bibinfo{booktitle}{\emph{CVPR}}. \bibinfo{pages}{3867--3876}.
\newblock


\bibitem[Lim et~al\mbox{.}(2017)]%
        {CVPRW2017EDSR}
\bibfield{author}{\bibinfo{person}{Bee Lim}, \bibinfo{person}{Sanghyun Son},
  \bibinfo{person}{Heewon Kim}, \bibinfo{person}{Seungjun Nah}, {and}
  \bibinfo{person}{Kyoung Mu~Lee}.} \bibinfo{year}{2017}\natexlab{}.
\newblock \showarticletitle{Enhanced deep residual networks for single image
  super-resolution}. In \bibinfo{booktitle}{\emph{CVPRW}}.
  \bibinfo{pages}{136--144}.
\newblock


\bibitem[Liu et~al\mbox{.}(2013)]%
        {CVPR2013filter1}
\bibfield{author}{\bibinfo{person}{Ming-Yu Liu}, \bibinfo{person}{Oncel Tuzel},
  {and} \bibinfo{person}{Yuichi Taguchi}.} \bibinfo{year}{2013}\natexlab{}.
\newblock \showarticletitle{Joint geodesic upsampling of depth images}. In
  \bibinfo{booktitle}{\emph{CVPR}}. \bibinfo{pages}{169--176}.
\newblock


\bibitem[Liu et~al\mbox{.}(2022a)]%
        {SPL2021DEAFNet}
\bibfield{author}{\bibinfo{person}{Peng Liu}, \bibinfo{person}{Zonghua Zhang},
  \bibinfo{person}{Zhaozong Meng}, {and} \bibinfo{person}{Nan Gao}.}
  \bibinfo{year}{2022}\natexlab{a}.
\newblock \showarticletitle{Deformable Enhancement and Adaptive Fusion for
  Depth Map Super-Resolution}.
\newblock \bibinfo{journal}{\emph{IEEE SPL}} (\bibinfo{year}{2022}),
  \bibinfo{pages}{204--208}.
\newblock


\bibitem[Liu et~al\mbox{.}(2022b)]%
        {NC2022PDRNet}
\bibfield{author}{\bibinfo{person}{Peng Liu}, \bibinfo{person}{Zonghua Zhang},
  \bibinfo{person}{Zhaozong Meng}, \bibinfo{person}{Nan Gao}, {and}
  \bibinfo{person}{Chao Wang}.} \bibinfo{year}{2022}\natexlab{b}.
\newblock \showarticletitle{PDR-Net: Progressive depth reconstruction network
  for color guided depth map super-resolution}.
\newblock \bibinfo{journal}{\emph{Neurocomputing}} (\bibinfo{year}{2022}).
\newblock


\bibitem[Lu et~al\mbox{.}(2014)]%
        {CVPR2014Lu}
\bibfield{author}{\bibinfo{person}{Si Lu}, \bibinfo{person}{Xiaofeng Ren},
  {and} \bibinfo{person}{Feng Liu}.} \bibinfo{year}{2014}\natexlab{}.
\newblock \showarticletitle{Depth enhancement via low-rank matrix completion}.
  In \bibinfo{booktitle}{\emph{CVPR}}. \bibinfo{pages}{3390--3397}.
\newblock


\bibitem[Lutio et~al\mbox{.}(2019)]%
        {ICCV2109PixTrasnform}
\bibfield{author}{\bibinfo{person}{Riccardo~de Lutio}, \bibinfo{person}{Stefano
  D'aronco}, \bibinfo{person}{Jan~Dirk Wegner}, {and} \bibinfo{person}{Konrad
  Schindler}.} \bibinfo{year}{2019}\natexlab{}.
\newblock \showarticletitle{Guided super-resolution as pixel-to-pixel
  transformation}. In \bibinfo{booktitle}{\emph{ICCV}}.
  \bibinfo{pages}{8829--8837}.
\newblock


\bibitem[Ning et~al\mbox{.}(2021)]%
        {NuerIPS2021UDL}
\bibfield{author}{\bibinfo{person}{Qian Ning}, \bibinfo{person}{Weisheng Dong},
  \bibinfo{person}{Xin Li}, \bibinfo{person}{Jinjian Wu}, {and}
  \bibinfo{person}{Guangming Shi}.} \bibinfo{year}{2021}\natexlab{}.
\newblock \showarticletitle{Uncertainty-Driven Loss for Single Image
  Super-Resolution}.
\newblock \bibinfo{journal}{\emph{NuerIPS}}.
\newblock


\bibitem[Pan et~al\mbox{.}(2019)]%
        {CVPR2019SVLRM}
\bibfield{author}{\bibinfo{person}{Jinshan Pan}, \bibinfo{person}{Jiangxin
  Dong}, \bibinfo{person}{Jimmy~S Ren}, \bibinfo{person}{Liang Lin},
  \bibinfo{person}{Jinhui Tang}, {and} \bibinfo{person}{Ming-Hsuan Yang}.}
  \bibinfo{year}{2019}\natexlab{}.
\newblock \showarticletitle{Spatially variant linear representation models for
  joint filtering}. In \bibinfo{booktitle}{\emph{CVPR}}.
  \bibinfo{pages}{1702--1711}.
\newblock


\bibitem[Park et~al\mbox{.}(2011)]%
        {ICCV2011Park}
\bibfield{author}{\bibinfo{person}{Jaesik Park}, \bibinfo{person}{Hyeongwoo
  Kim}, \bibinfo{person}{Yu-Wing Tai}, \bibinfo{person}{Michael~S Brown}, {and}
  \bibinfo{person}{Inso Kweon}.} \bibinfo{year}{2011}\natexlab{}.
\newblock \showarticletitle{High quality depth map upsampling for 3d-tof
  cameras}. In \bibinfo{booktitle}{\emph{ICCV}}. \bibinfo{pages}{1623--1630}.
\newblock


\bibitem[Poggi et~al\mbox{.}(2020)]%
        {CVPR2020UncertaintyMDE}
\bibfield{author}{\bibinfo{person}{Matteo Poggi}, \bibinfo{person}{Filippo
  Aleotti}, \bibinfo{person}{Fabio Tosi}, {and} \bibinfo{person}{Stefano
  Mattoccia}.} \bibinfo{year}{2020}\natexlab{}.
\newblock \showarticletitle{On the uncertainty of self-supervised monocular
  depth estimation}. In \bibinfo{booktitle}{\emph{CVPR}}.
  \bibinfo{pages}{3227--3237}.
\newblock


\bibitem[Riegler et~al\mbox{.}(2016)]%
        {ECCV2016atgvnet}
\bibfield{author}{\bibinfo{person}{Gernot Riegler}, \bibinfo{person}{Matthias
  R{\"u}ther}, {and} \bibinfo{person}{Horst Bischof}.}
  \bibinfo{year}{2016}\natexlab{}.
\newblock \showarticletitle{Atgv-net: Accurate depth super-resolution}. In
  \bibinfo{booktitle}{\emph{ECCV}}. \bibinfo{pages}{268--284}.
\newblock


\bibitem[Scharstein and Pal(2007)]%
        {CVPR2007Middlebury2005}
\bibfield{author}{\bibinfo{person}{Daniel Scharstein} {and}
  \bibinfo{person}{Chris Pal}.} \bibinfo{year}{2007}\natexlab{}.
\newblock \showarticletitle{Learning conditional random fields for stereo}. In
  \bibinfo{booktitle}{\emph{CVPR}}. \bibinfo{pages}{1--8}.
\newblock


\bibitem[Silberman et~al\mbox{.}(2012)]%
        {ECCV2012NYUv2}
\bibfield{author}{\bibinfo{person}{Nathan Silberman}, \bibinfo{person}{Derek
  Hoiem}, \bibinfo{person}{Pushmeet Kohli}, {and} \bibinfo{person}{Rob
  Fergus}.} \bibinfo{year}{2012}\natexlab{}.
\newblock \showarticletitle{Indoor segmentation and support inference from rgbd
  images}. In \bibinfo{booktitle}{\emph{ECCV}}. \bibinfo{pages}{746--760}.
\newblock


\bibitem[Su et~al\mbox{.}(2019)]%
        {CVPR2019PAC}
\bibfield{author}{\bibinfo{person}{Hang Su}, \bibinfo{person}{Varun Jampani},
  \bibinfo{person}{Deqing Sun}, \bibinfo{person}{Orazio Gallo},
  \bibinfo{person}{Erik Learned-Miller}, {and} \bibinfo{person}{Jan Kautz}.}
  \bibinfo{year}{2019}\natexlab{}.
\newblock \showarticletitle{Pixel-adaptive convolutional neural networks}. In
  \bibinfo{booktitle}{\emph{CVPR}}. \bibinfo{pages}{11166--11175}.
\newblock


\bibitem[Sun et~al\mbox{.}(2021)]%
        {CVPR2021CTKT}
\bibfield{author}{\bibinfo{person}{Baoli Sun}, \bibinfo{person}{Xinchen Ye},
  \bibinfo{person}{Baopu Li}, \bibinfo{person}{Haojie Li},
  \bibinfo{person}{Zhihui Wang}, {and} \bibinfo{person}{Rui Xu}.}
  \bibinfo{year}{2021}\natexlab{}.
\newblock \showarticletitle{Learning scene structure guidance via cross-task
  knowledge transfer for single depth super-resolution}. In
  \bibinfo{booktitle}{\emph{CVPR}}. \bibinfo{pages}{7792--7801}.
\newblock


\bibitem[Tang et~al\mbox{.}(2021a)]%
        {MM21JIIF}
\bibfield{author}{\bibinfo{person}{Jiaxiang Tang}, \bibinfo{person}{Xiaokang
  Chen}, {and} \bibinfo{person}{Gang Zeng}.} \bibinfo{year}{2021}\natexlab{a}.
\newblock \showarticletitle{Joint implicit image function for guided depth
  super-resolution}. In \bibinfo{booktitle}{\emph{ACM MM}}.
  \bibinfo{pages}{4390--4399}.
\newblock


\bibitem[Tang et~al\mbox{.}(2021b)]%
        {MM2021bridgenet}
\bibfield{author}{\bibinfo{person}{Qi Tang}, \bibinfo{person}{Runmin Cong},
  \bibinfo{person}{Ronghui Sheng}, \bibinfo{person}{Lingzhi He},
  \bibinfo{person}{Dan Zhang}, \bibinfo{person}{Yao Zhao}, {and}
  \bibinfo{person}{Sam Kwong}.} \bibinfo{year}{2021}\natexlab{b}.
\newblock \showarticletitle{BridgeNet: A Joint Learning Network of Depth Map
  Super-Resolution and Monocular Depth Estimation}. In
  \bibinfo{booktitle}{\emph{ACM MM}}. \bibinfo{pages}{2148--2157}.
\newblock


\bibitem[Van~Gansbeke et~al\mbox{.}(2019)]%
        {MVA2019uncertainty}
\bibfield{author}{\bibinfo{person}{Wouter Van~Gansbeke}, \bibinfo{person}{Davy
  Neven}, \bibinfo{person}{Bert De~Brabandere}, {and} \bibinfo{person}{Luc
  Van~Gool}.} \bibinfo{year}{2019}\natexlab{}.
\newblock \showarticletitle{Sparse and noisy lidar completion with rgb guidance
  and uncertainty}. In \bibinfo{booktitle}{\emph{MVA}}. \bibinfo{pages}{1--6}.
\newblock


\bibitem[Wang et~al\mbox{.}(2020a)]%
        {TPAMI2020survey}
\bibfield{author}{\bibinfo{person}{Zhihao Wang}, \bibinfo{person}{Jian Chen},
  {and} \bibinfo{person}{Steven~CH Hoi}.} \bibinfo{year}{2020}\natexlab{a}.
\newblock \showarticletitle{Deep learning for image super-resolution: A
  survey}.
\newblock \bibinfo{journal}{\emph{IEEE TPAMI}} (\bibinfo{year}{2020}),
  \bibinfo{pages}{3365--3387}.
\newblock


\bibitem[Wang et~al\mbox{.}(2020b)]%
        {PR2020DSRN}
\bibfield{author}{\bibinfo{person}{Zhihui Wang}, \bibinfo{person}{Xinchen Ye},
  \bibinfo{person}{Baoli Sun}, \bibinfo{person}{Jingyu Yang},
  \bibinfo{person}{Rui Xu}, {and} \bibinfo{person}{Haojie Li}.}
  \bibinfo{year}{2020}\natexlab{b}.
\newblock \showarticletitle{Depth upsampling based on deep edge-aware
  learning}.
\newblock \bibinfo{journal}{\emph{PR}} (\bibinfo{year}{2020}),
  \bibinfo{pages}{107274}.
\newblock


\bibitem[Wen et~al\mbox{.}(2018)]%
        {TIP2018CCFN}
\bibfield{author}{\bibinfo{person}{Yang Wen}, \bibinfo{person}{Bin Sheng},
  \bibinfo{person}{Ping Li}, \bibinfo{person}{Weiyao Lin}, {and}
  \bibinfo{person}{David~Dagan Feng}.} \bibinfo{year}{2018}\natexlab{}.
\newblock \showarticletitle{Deep color guided coarse-to-fine convolutional
  network cascade for depth image super-resolution}.
\newblock \bibinfo{journal}{\emph{IEEE TIP}} (\bibinfo{year}{2018}),
  \bibinfo{pages}{994--1006}.
\newblock


\bibitem[Woo et~al\mbox{.}(2018)]%
        {ECCV2018CBAM}
\bibfield{author}{\bibinfo{person}{Sanghyun Woo}, \bibinfo{person}{Jongchan
  Park}, \bibinfo{person}{Joon-Young Lee}, {and} \bibinfo{person}{In~So
  Kweon}.} \bibinfo{year}{2018}\natexlab{}.
\newblock \showarticletitle{Cbam: Convolutional block attention module}. In
  \bibinfo{booktitle}{\emph{ECCV}}. \bibinfo{pages}{3--19}.
\newblock


\bibitem[Wright(1999)]%
        {TNN1999BNN}
\bibfield{author}{\bibinfo{person}{WA Wright}.}
  \bibinfo{year}{1999}\natexlab{}.
\newblock \showarticletitle{Bayesian approach to neural-network modeling with
  input uncertainty}.
\newblock \bibinfo{journal}{\emph{IEEE TNN}} (\bibinfo{year}{1999}),
  \bibinfo{pages}{1261--1270}.
\newblock


\bibitem[Xie et~al\mbox{.}(2015)]%
        {TIP2015MRF}
\bibfield{author}{\bibinfo{person}{Jun Xie}, \bibinfo{person}{Rogerio~Schmidt
  Feris}, {and} \bibinfo{person}{Ming-Ting Sun}.}
  \bibinfo{year}{2015}\natexlab{}.
\newblock \showarticletitle{Edge-guided single depth image super resolution}.
\newblock \bibinfo{journal}{\emph{IEEE TIP}} (\bibinfo{year}{2015}),
  \bibinfo{pages}{428--438}.
\newblock


\bibitem[Xu et~al\mbox{.}(2020)]%
        {CVPR2020uncertainty}
\bibfield{author}{\bibinfo{person}{Ke Xu}, \bibinfo{person}{Xin Yang},
  \bibinfo{person}{Baocai Yin}, {and} \bibinfo{person}{Rynson~WH Lau}.}
  \bibinfo{year}{2020}\natexlab{}.
\newblock \showarticletitle{Learning to restore low-light images via
  decomposition-and-enhancement}. In \bibinfo{booktitle}{\emph{CVPR}}.
  \bibinfo{pages}{2281--2290}.
\newblock


\bibitem[Ye et~al\mbox{.}(2021)]%
        {ICCV21CA}
\bibfield{author}{\bibinfo{person}{Mang Ye}, \bibinfo{person}{Weijian Ruan},
  \bibinfo{person}{Bo Du}, {and} \bibinfo{person}{Mike~Zheng Shou}.}
  \bibinfo{year}{2021}\natexlab{}.
\newblock \showarticletitle{Channel augmented joint learning for
  visible-infrared recognition}. In \bibinfo{booktitle}{\emph{ICCV}}.
  \bibinfo{pages}{13567--13576}.
\newblock


\bibitem[Ye et~al\mbox{.}(2020a)]%
        {TPAMI2020yemang}
\bibfield{author}{\bibinfo{person}{Mang Ye}, \bibinfo{person}{Jianbing Shen},
  \bibinfo{person}{Xu Zhang}, \bibinfo{person}{Pong~C Yuen}, {and}
  \bibinfo{person}{Shih-Fu Chang}.} \bibinfo{year}{2020}\natexlab{a}.
\newblock \showarticletitle{Augmentation invariant and instance spreading
  feature for softmax embedding}.
\newblock \bibinfo{journal}{\emph{IEEE TMAPI}} (\bibinfo{year}{2020}).
\newblock


\bibitem[Ye et~al\mbox{.}(2020b)]%
        {MM2020SCSN}
\bibfield{author}{\bibinfo{person}{Xinchen Ye}, \bibinfo{person}{Baoli Sun},
  \bibinfo{person}{Zhihui Wang}, \bibinfo{person}{Jingyu Yang},
  \bibinfo{person}{Rui Xu}, \bibinfo{person}{Haojie Li}, {and}
  \bibinfo{person}{Baopu Li}.} \bibinfo{year}{2020}\natexlab{b}.
\newblock \showarticletitle{Depth super-resolution via deep controllable
  slicing network}. In \bibinfo{booktitle}{\emph{ACM MM}}.
  \bibinfo{pages}{1809--1818}.
\newblock


\bibitem[Ye et~al\mbox{.}(2020c)]%
        {TIP2020pmbanet}
\bibfield{author}{\bibinfo{person}{Xinchen Ye}, \bibinfo{person}{Baoli Sun},
  \bibinfo{person}{Zhihui Wang}, \bibinfo{person}{Jingyu Yang},
  \bibinfo{person}{Rui Xu}, \bibinfo{person}{Haojie Li}, {and}
  \bibinfo{person}{Baopu Li}.} \bibinfo{year}{2020}\natexlab{c}.
\newblock \showarticletitle{Pmbanet: Progressive multi-branch aggregation
  network for scene depth super-resolution}.
\newblock \bibinfo{journal}{\emph{IEEE TIP}} (\bibinfo{year}{2020}),
  \bibinfo{pages}{7427--7442}.
\newblock


\bibitem[Zagoruyko and Komodakis(2017)]%
        {ICLR2017attention}
\bibfield{author}{\bibinfo{person}{Sergey Zagoruyko} {and}
  \bibinfo{person}{Nikos Komodakis}.} \bibinfo{year}{2017}\natexlab{}.
\newblock \showarticletitle{Paying More Attention to Attention: Improving the
  Performance of Convolutional Neural Networks via Attention Transfer}. In
  \bibinfo{booktitle}{\emph{ICLR}}.
\newblock


\bibitem[Zhang et~al\mbox{.}(2018)]%
        {ECCV2018RCAN}
\bibfield{author}{\bibinfo{person}{Yulun Zhang}, \bibinfo{person}{Kunpeng Li},
  \bibinfo{person}{Kai Li}, \bibinfo{person}{Lichen Wang},
  \bibinfo{person}{Bineng Zhong}, {and} \bibinfo{person}{Yun Fu}.}
  \bibinfo{year}{2018}\natexlab{}.
\newblock \showarticletitle{Image super-resolution using very deep residual
  channel attention networks}. In \bibinfo{booktitle}{\emph{ECCV}}.
  \bibinfo{pages}{286--301}.
\newblock


\bibitem[Zhao et~al\mbox{.}(2019)]%
        {PR2019CDcGAN}
\bibfield{author}{\bibinfo{person}{Lijun Zhao}, \bibinfo{person}{Huihui Bai},
  \bibinfo{person}{Jie Liang}, \bibinfo{person}{Bing Zeng},
  \bibinfo{person}{Anhong Wang}, {and} \bibinfo{person}{Yao Zhao}.}
  \bibinfo{year}{2019}\natexlab{}.
\newblock \showarticletitle{Simultaneous color-depth super-resolution with
  conditional generative adversarial networks}.
\newblock \bibinfo{journal}{\emph{PR}} (\bibinfo{year}{2019}),
  \bibinfo{pages}{356--369}.
\newblock


\bibitem[Zhu et~al\mbox{.}(2022)]%
        {AAAI2022uncertainty}
\bibfield{author}{\bibinfo{person}{Yufan Zhu}, \bibinfo{person}{Weisheng Dong},
  \bibinfo{person}{Leida Li}, \bibinfo{person}{Jinjian Wu},
  \bibinfo{person}{Xin Li}, {and} \bibinfo{person}{Guangming Shi}.}
  \bibinfo{year}{2022}\natexlab{}.
\newblock \showarticletitle{Robust depth completion with uncertainty-driven
  loss functions}. In \bibinfo{booktitle}{\emph{AAAI}}.
  \bibinfo{pages}{3626--3634}.
\newblock


\bibitem[Zuo et~al\mbox{.}(2019)]%
        {TCSVT2019MFR}
\bibfield{author}{\bibinfo{person}{Yifan Zuo}, \bibinfo{person}{Qiang Wu},
  \bibinfo{person}{Yuming Fang}, \bibinfo{person}{Ping An},
  \bibinfo{person}{Liqin Huang}, {and} \bibinfo{person}{Zhifeng Chen}.}
  \bibinfo{year}{2019}\natexlab{}.
\newblock \showarticletitle{Multi-scale frequency reconstruction for guided
  depth map super-resolution via deep residual network}.
\newblock \bibinfo{journal}{\emph{IEEE TCSVT}} (\bibinfo{year}{2019}),
  \bibinfo{pages}{297--306}.
\newblock


\end{thebibliography}

\end{document}